\documentclass[10pt,twocolumn,letterpaper]{article}

\usepackage[pagenumbers]{cvpr}

\usepackage{cuted} 

\usepackage{currfile} 

\usepackage{caption} 

\usepackage{microtype}

\usepackage{multirow}
\usepackage{colortbl}
\definecolor{cvprblue}{rgb}{0.21,0.49,0.74}
\usepackage[breaklinks,colorlinks,allcolors=cvprblue]{hyperref}

\title{HS-OVSeg: Multi-Granularity Open-Vocabulary Semantic Segmentation in Remote Sensing via Heterogeneous Supervision}

\author{Wei Huang$^{1,3}$ \quad Chenying Liu$^{1,3}$ \quad Xudong Kang$^{2}$ \quad Xiao Xiang Zhu$^{1,3}$\\[2pt]
$^{1}$Technical University of Munich (TUM), Germany\\
$^{2}$Hunan University, China\\
$^{3}$Munich Center for Machine Learning (MCML), Germany}

\hypersetup{pdfauthor={Wei Huang, Chenying Liu, Xudong Kang, Xiao Xiang Zhu},
 pdftitle={HS-OVSeg: Multi-Granularity Open-Vocabulary Semantic Segmentation in Remote Sensing via Heterogeneous Supervision}}

\begin{document}
\maketitle

\begin{strip}
    \vspace{-1.2em}
    \centering
    \begin{minipage}{1.0\textwidth}
        \centering
        \includegraphics[
            width=\linewidth,
            height=0.38\textheight,
            keepaspectratio
        ]{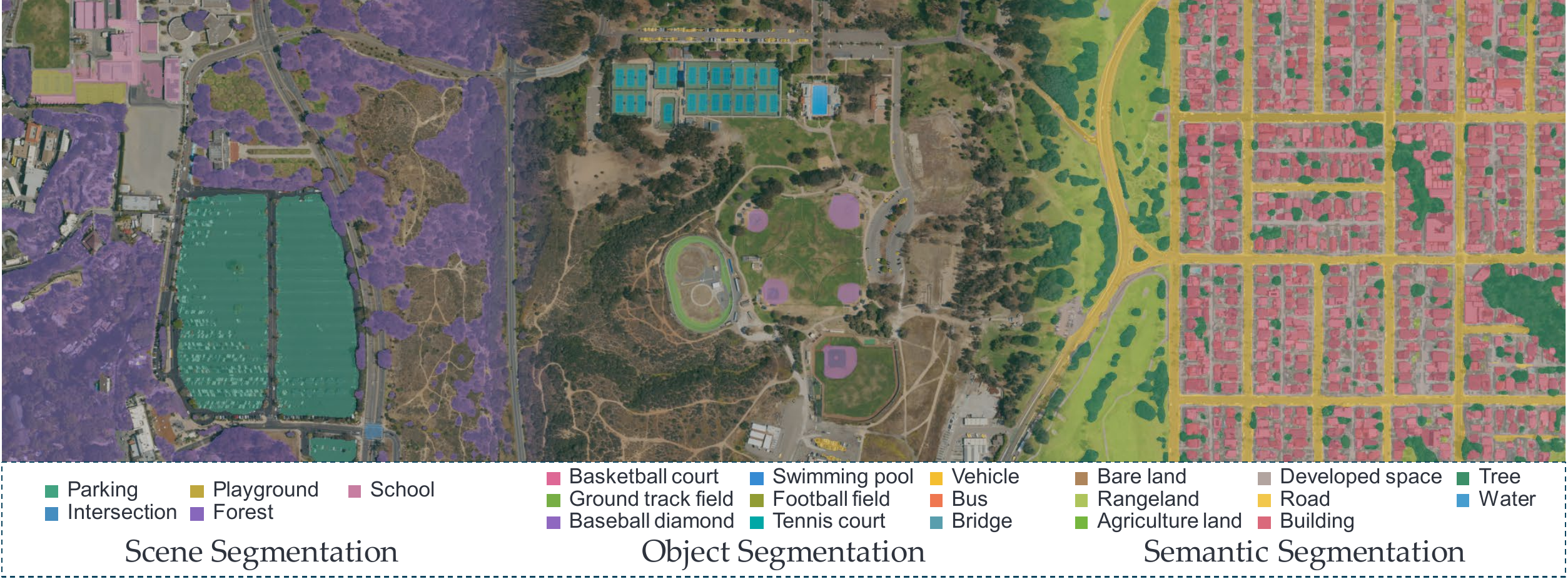}
        \vspace{-1.1em}
        \captionsetup{hypcap=false}
        \captionof{figure}{Multi-granularity open-vocabulary semantic segmentation with HS-OVSeg.}
        \captionsetup{hypcap=true}
        \label{fig:multi_granularity_prediction}
    \end{minipage}
    \par\vspace{7pt}
\end{strip}

\begin{abstract}
Open-vocabulary semantic segmentation (OVSS) enables flexible, text-driven mapping of remote sensing (RS) imagery beyond predefined categories. Despite advances in segmentation foundation models, RS-OVSS remains constrained by the scarcity of large-scale, semantically diverse datasets with accurate pixel-level annotations. Meanwhile, existing RS datasets contain abundant scene- and object-level annotations that offer valuable semantic information but lack precise pixel-wise supervision. To exploit these complementary sources, we propose \textbf{HS-OVSeg}, a unified training framework that integrates heterogeneous scene-, object-, and pixel-level supervision for RS-OVSS. Specifically, we generate pixel-wise pseudo-labels using SAM3, conditioned on scene-level class names and object-level class names with bounding boxes. Combined with ground-truth semantic masks, these pseudo-labels enable joint training across all three annotation granularities through LoRA-based fine-tuning of SAM3. Our unified training spans 20 RS datasets, including 3 scene classification, 4 object detection, and 13 semantic segmentation datasets, with 1,000 images from the 7 classification and detection datasets manually annotated for pixel-level evaluation. HS-OVSeg substantially outperforms existing RS-OVSS models across both the 20 training-source and 5 held-out datasets, demonstrating strong multi-granularity segmentation capabilities under heterogeneous supervision. The code will be available at \url{https://github.com/zhu-xlab/HS-OVSeg}.
\end{abstract}

\section{Introduction}
\label{sec:introduction}

Remote sensing (RS) imagery captures rich semantics at multiple levels, from pixel-level land-cover categories (e.g., vegetation) and object-level entities (e.g., cars) to scene-level concepts (e.g., airports). Diverse RS applications require recognition beyond predefined categories, which conventional semantic segmentation models struggle to accommodate. Open-vocabulary semantic segmentation (OVSS) addresses this limitation through flexible, text-driven semantic mapping \cite{cat_seg_2024,proxyclip_2024}. Advances in vision-language models (VLMs), such as CLIP \cite{clip_2021}, dense vision-language segmentation methods \cite{maskclip_2022,sclip_2024,cat_seg_2024}, and segmentation foundation models such as SAM3 \cite{sam3_2026} have further advanced open-vocabulary recognition and pixel-level localization.

\begin{figure*}[ht]
    \centering
    \includegraphics[width=\textwidth]{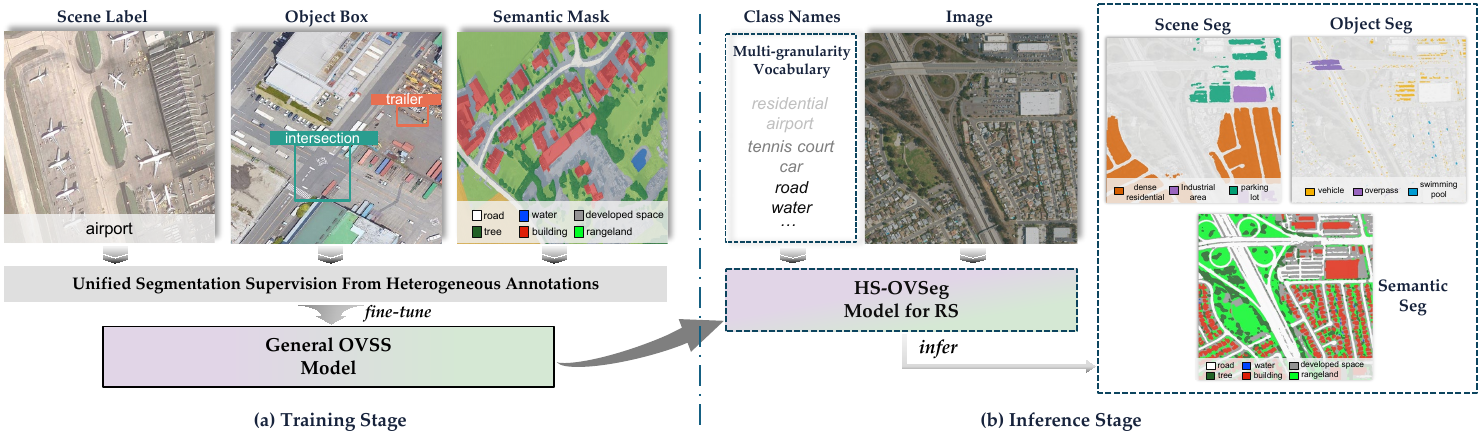}
    \vspace{-0.5cm}
    \caption{
        Illustration of heterogeneous supervision for RS-OVSS, i.e., HS-OVSeg.
        (a) Training: Scene labels, classes with object boxes, and semantic masks provide complementary supervision at different spatial granularities and semantic scales, offering potential for unified OVSS fine-tuning.
        (b) Inference: Given an image and category names, the model segments pixel-, object-, and scene-level concepts.
    }
    \label{fig:heterogeneous_sources}
    \vspace{-0.2cm}
\end{figure*}

RS-OVSS has also advanced through RS-aware vision-language representations and segmentation \cite{gsnet_2025,rskt_seg_2026}, training-free dense prediction \cite{segearth_ov_2025,rsclip_2026,ldseg_2026,reattnclip_2026}, and foundation model-based segmentation \cite{segearth_ov3_2025}. However, progress remains constrained by the scarcity of large-scale, semantically diverse datasets with accurate pixel-level annotations. Existing RS segmentation datasets primarily cover land-use/land-cover categories, with limited representation of diverse object- and scene-level concepts. Obtaining comprehensive pixel-level annotations across these semantic levels is prohibitively expensive.

An opportunity lies in the heterogeneous annotations available across RS datasets. As illustrated in Fig.~\ref{fig:heterogeneous_sources}, scene labels indicate overall semantic concepts, object annotations provide class names and bounding boxes, and pixel annotations offer precise semantic masks. Coarser annotations often cover diverse scene and object concepts underrepresented in segmentation datasets. However, their complementary semantic coverage comes with different spatial supervision granularities, making it challenging to integrate them into a unified RS-OVSS learning objective.

To address this challenge, we propose \textbf{HS-OVSeg}, a unified framework that jointly exploits scene-, object-, and pixel-level annotations for RS-OVSS. The key idea is to unify different spatial supervision granularities into a common pixel-level learning signal while leveraging complementary semantic coverage. Specifically, scene-level labels provide semantic cues for pixel-level pseudo-label generation, while object-level annotations offer geometric guidance through bounding boxes. These pseudo-labels are combined with ground-truth (GT) semantic masks under a shared objective for concept presence and pixel-level segmentation. Built upon SAM3, HS-OVSeg optimizes shared low-rank adapters~\cite{lora_2022} across all three annotation levels while keeping pretrained weights frozen. At inference, the adapted model segments pixel-, object-, and scene-level concepts using only images and category names, as illustrated in Fig.~\ref{fig:heterogeneous_sources}.

We train HS-OVSeg on 20 RS datasets, comprising 3 scene classification, 4 object detection, and 13 semantic segmentation datasets, with 1,000 images from the 7 classification and detection datasets manually annotated for pixel-level evaluation. HS-OVSeg substantially outperforms existing RS-OVSS models across these datasets and demonstrates strong generalization on 5 held-out datasets.

Our main contributions are summarized as follows:
\begin{itemize}
    \item We propose \textbf{HS-OVSeg}, a unified segmentation framework that jointly learns from semantically complementary scene-, object-, and pixel-level annotations despite their different spatial granularities, extending RS-OVSS training beyond pixel-annotated datasets.

    \item We develop an annotation-guided pseudo-labeling strategy that uses semantic cues, geometric guidance, and pretrained spatial priors to convert coarse annotations into a common pixel-level supervision signal for joint learning.

    \item HS-OVSeg substantially outperforms existing RS-OVSS methods across 20 training-source datasets spanning all three annotation granularities and demonstrates strong cross-dataset generalization on 5 held-out datasets.
\end{itemize}

\section{Related Work}
\label{sec:related_work}

\subsection{Open-Vocabulary Semantic Segmentation}

OVSS extends semantic segmentation beyond predefined categories by leveraging VLMs, particularly CLIP \cite{clip_2021}. Existing methods generally follow training-based and training-free paradigms. Training-based approaches improve dense visual-semantic alignment through mask-aware adaptation in OVSeg \cite{ovseg_2023}, lightweight adapters in SAN \cite{san_2023}, cost aggregation in CAT-Seg \cite{cat_seg_2024}, and embedding balancing in EBSeg \cite{ebseg_2024}. Training-free approaches exploit pretrained representations through dense label extraction in MaskCLIP \cite{maskclip_2022}, attention and feature refinement in SCLIP \cite{sclip_2024}, ClearCLIP \cite{lan2024clearclip}, and NACLIP \cite{naclip_2025}, vision foundation model-guided attention in ProxyCLIP \cite{proxyclip_2024}, and patch correlation reconstruction in CorrCLIP \cite{corrclip_2025}. GEM \cite{gem_2024} further explores zero-shot visual grounding, while Trident \cite{trident_2025} integrates multiple foundation models for training-free segmentation. Recently, SAM3 \cite{sam3_2026} has advanced concept-conditioned segmentation through text prompts.

In RS, RemoteCLIP \cite{remoteclip_2024}, GeoRSCLIP \cite{georsclip_2024}, and SkySense-O \cite{skysense_o_2025} have advanced open-vocabulary visual understanding and semantic localization. For RS-OVSS, GSNet \cite{gsnet_2025} combines RS-specific and general vision-language features, while RSKT-Seg \cite{rskt_seg_2026} introduces efficient cost aggregation and RS knowledge transfer. Training-free approaches, including SegEarth-OV \cite{segearth_ov_2025}, RSCLIP \cite{rsclip_2026}, LDSeg \cite{ldseg_2026}, and ReAttnCLIP \cite{reattnclip_2026}, enhance dense semantic prediction through feature, attention, and spatial refinement. More recently, SegEarth-OV3 \cite{segearth_ov3_2025} explores SAM3-based segmentation through dual-head prediction fusion and category filtering.

\subsection{Multi-Task and Heterogeneous Supervision}

Weakly supervised approaches have explored learning segmentation from image-text supervision, such as text-guided region grouping in GroupViT \cite{groupvit_2022} and region-word alignment in OpenSeg \cite{openseg_2022}. Beyond weak supervision, unified frameworks have integrated diverse segmentation and vision-language tasks through generalized decoding in X-Decoder \cite{xdecoder_2023}, multimodal prompting in SEEM \cite{seem_2023}, and multi-task segmentation architectures in OMG-Seg \cite{omgseg_2024} and QueryMeldNet \cite{querymeldnet_2025}. More closely related, OpenSeeD \cite{openseed_2023} jointly learns from detection and segmentation by bridging vocabulary and annotation granularity gaps. In RS, RemoteCLIP \cite{remoteclip_2024} converts detection and segmentation annotations into image-text training pairs.

Unlike existing approaches that focus on multi-task unification or limited annotation types, HS-OVSeg jointly exploits scene-, object-, and pixel-level supervision under a unified RS-OVSS objective. By integrating these heterogeneous annotations, HS-OVSeg extends training beyond conventional pixel-annotated datasets, enabling broader semantic coverage across diverse scene and object concepts.

\section{HS-OVSeg}
\label{sec:method}

\begin{figure*}[t]
    \centering
    \includegraphics[width=\textwidth]{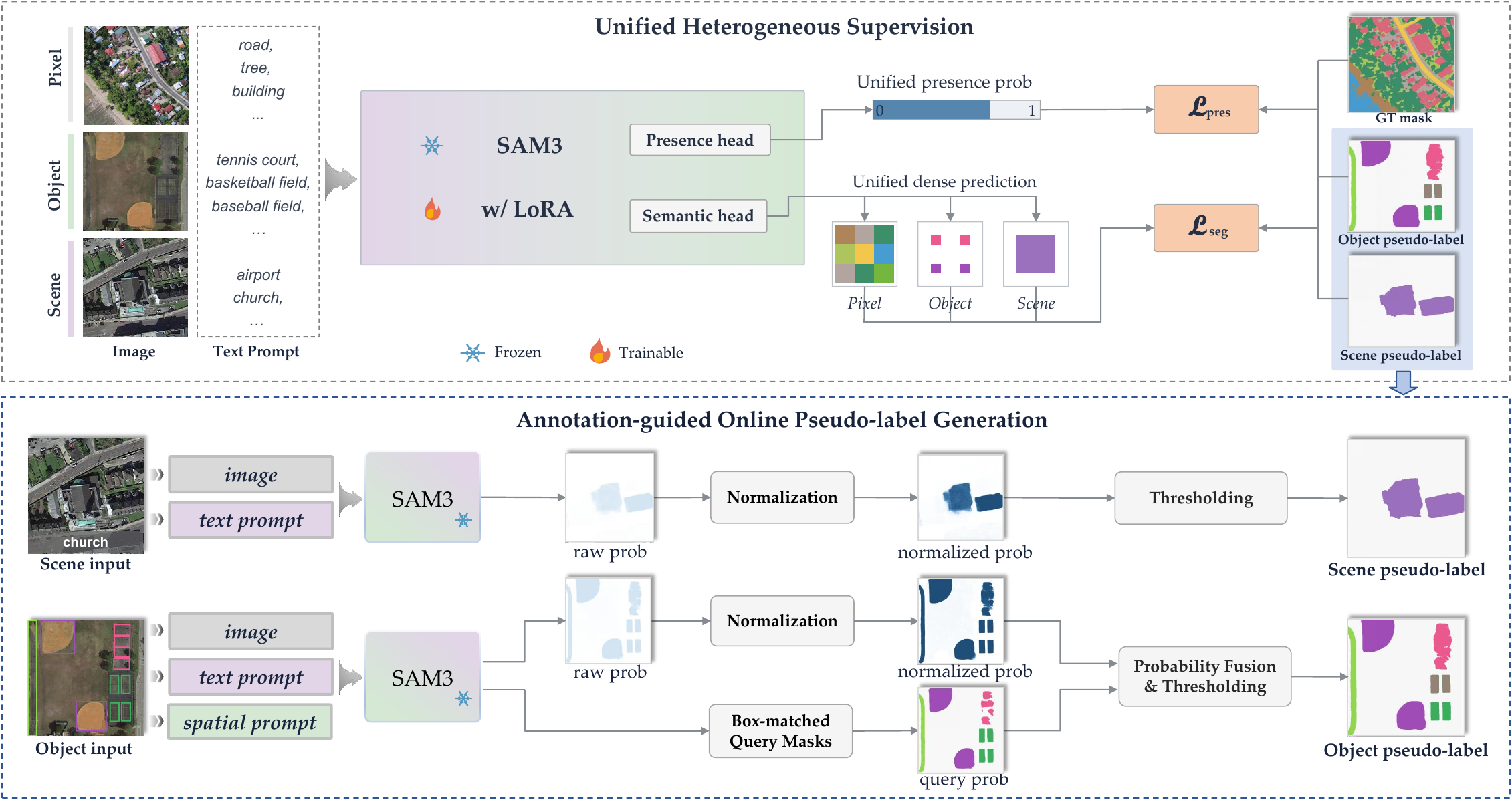}
    \vspace{-0.5cm}
    \caption{Overview of HS-OVSeg. Top: unified LoRA adaptation of SAM3 using shared presence and segmentation objectives across scene-, object-, and pixel-level supervision. Bottom: annotation-guided online pseudo-label generation for scene and object annotations.}
    \label{fig:hs_ovseg}
    \vspace{-0.1cm}
\end{figure*}

As shown in Fig.~\ref{fig:hs_ovseg}, HS-OVSeg adopts a teacher--student framework to unify heterogeneous scene-, object-, and pixel-level supervision for RS-OVSS. The frozen pretrained SAM3 serves as the teacher, transforming scene- and object-level annotations into pixel-level pseudo-labels. The student adapts SAM3 through low-rank adaptation (LoRA)~\cite{lora_2022}, jointly learning from these pseudo-labels and GT semantic masks under a unified supervision objective.

\subsection{Preliminaries of SAM3 for HS-OVSeg}
\label{sec:sam3_preliminaries}

We introduce only the SAM3~\cite{sam3_2026} components used in HS-OVSeg. Its Perception Encoder (PE) encodes the input image and category text, while a multimodal fusion encoder produces class-conditioned visual features; a query-based decoder predicts class-matching scores, bounding boxes, and instance masks; a presence head estimates whether the queried class occurs in the image, and a semantic head produces its dense binary segmentation map; for object-level pseudo-label generation, a geometry encoder incorporates bounding boxes as spatial prompts.

For an image $I$, a category name $c$, and optional box prompts $\mathcal{B}$, we denote the relevant probability outputs as
\begin{equation}
\mathcal{F}_{\theta}(I,c,\mathcal{B})
=
\left(
a_c,\,
S_c,\,
\left\{q_{c,j},M_{c,j},b_{c,j}\right\}_{j=1}^{N}
\right),
\label{eq:sam3_outputs}
\end{equation}
where $a_c$ is the presence probability, $S_c$ is the dense semantic probability map, and each query $j$ predicts a class-matching probability $q_{c,j}$, an instance-mask probability map $M_{c,j}$, and a bounding box $b_{c,j}$. All probability outputs are obtained by applying sigmoid activation to their corresponding logits. 

\subsection{Unified Heterogeneous Supervision}
\label{sec:unified_adaptation}

\paragraph{Parameter-efficient adaptation.}
We employ LoRA to adapt SAM3 while keeping its pretrained parameters frozen. For an eligible pretrained weight $W_0\in\mathbb{R}^{d_{\mathrm{out}}\times d_{\mathrm{in}}}$, its effective weight is
\begin{equation}
W = W_0 + BA,
\qquad
A\in\mathbb{R}^{r\times d_{\mathrm{in}}},
\quad
B\in\mathbb{R}^{d_{\mathrm{out}}\times r},
\label{eq:lora}
\end{equation}
where $r$ is the adaptation rank, set to $8$ in our experiments. For convolutional kernels, dimensions following the output-channel dimension are flattened for the low-rank update, which is then reshaped back to the original kernel shape.

In HS-OVSeg, the presence and fused segmentation losses optimize low-rank adapters along the PE image and text encoders, multimodal fusion encoder, query-based decoder, and presence, semantic, and query prediction pathways. The pixel-wise fusion module introduced below is jointly optimized. Each eligible weight has its own low-rank factors, and the same set of adapters is shared across scene-, object-, and pixel-level supervision.

\paragraph{A common pixel-level target.}
Let $I\in\mathbb{R}^{3\times H\times W}$ be a training image with annotation granularity $\ell\in\{\mathrm{scene},\mathrm{object},\mathrm{pixel}\}$. Let $\mathcal{C}_I$ denote the set of category prompts sampled for image $I$, including both positive and negative categories, and let $K=|\mathcal{C}_I|$ be the number of sampled categories. Positive categories are prioritized, with remaining slots sampled from annotation-absent categories in the same dataset vocabulary. For each category $c\in\mathcal{C}_I$, we construct a binary pixel target $Y_c^{(\ell)}\in\{0,1\}^{H\times W}$:
\begin{equation}
Y_c^{(\ell)}
=
\begin{cases}
\widehat{Y}^{\mathrm{scene}}_c,
& \ell=\mathrm{scene},\\
\widehat{Y}^{\mathrm{object}}_c,
& \ell=\mathrm{object},\\
\mathbf{1}[Y^{\mathrm{pixel}}=c],
& \ell=\mathrm{pixel},
\end{cases}
\label{eq:unified_targets}
\end{equation}
where $\widehat{Y}^{\mathrm{scene}}_c,\widehat{Y}^{\mathrm{object}}_c\in\{0,1\}^{H\times W}$ are online pseudo-labels described in Sec.~\ref{sec:pseudo_labels}, and $Y^{\mathrm{pixel}}\in\mathcal{V}_d^{H\times W}$ is the GT semantic label map, with $\mathcal{V}_d$ denoting the dataset-specific set of category IDs. Invalid or unlabeled pixels are ignored when computing the losses; object supervision additionally excludes regions of omitted same-category boxes.

\paragraph{Unified prediction and optimization.}
For each category $c$, we aggregate the query mask probabilities into a category-wise map:
\begin{equation}
Q_c=\mathcal{R}\!\left(
\max_{j\in\mathcal{J}_c}
(\kappa_{c,j}q_{c,j})M_{c,j}
\right),
\label{eq:student_query_aggregation}
\end{equation}
where $\mathcal{J}_c$ contains the eight queries with the highest stability-weighted scores $\kappa_{c,j}q_{c,j}$, $\mathcal{R}$ denotes bilinear resizing, and $\kappa_{c,j}$ measures mask stability under small variations in the binarization threshold (Supplementary Sec.~\ref{sec:supp_query_stability}).

We calibrate the semantic and aggregated query logits as $\widetilde{s}_c=\operatorname{logit}(S_c)/T_S+\beta_S$ and $\widetilde{q}_c=\operatorname{logit}(Q_c)/T_Q+\beta_Q$, where $T_S,T_Q>0$ are learned temperatures controlling the respective logit scales, and $\beta_S,\beta_Q$ are learned biases. The calibrated logits are fused as
\begin{equation}
\begin{aligned}
Z_c&=\tfrac{1}{2}(\widetilde{s}_c+\widetilde{q}_c)
+g_{\phi}([\widetilde{s}_c,\widetilde{q}_c]),\\
F_c&=\sigma(Z_c).
\end{aligned}
\label{eq:pixel_fusion}
\end{equation}
The local fusion module $g_{\phi}$ comprises two $3\times3$ convolutions ($2\!\to\!16\!\to\!1$) separated by GELU, yielding a $5\times5$ receptive field. 

Stacking the fused logits $Z_c$, probabilities $F_c=\sigma(Z_c)$, and binary targets $Y_c^{(\ell)}$ over the $K$ sampled categories yields $Z\in\mathbb{R}^{K\times H\times W}$, $F\in[0,1]^{K\times H\times W}$, and $Y\in\{0,1\}^{K\times H\times W}$, respectively. Across all three annotation granularities, we apply the same segmentation objective:
\begin{equation}
\mathcal{L}_{\mathrm{seg}}
=
\mathcal{L}_{\mathrm{BCE}}(F,Y)
+
\mathcal{L}_{\mathrm{Dice}}(F,Y)
+
\mathcal{L}_{\mathrm{CE}}(Z,Y).
\label{eq:segmentation_loss}
\end{equation}
BCE and Dice supervise valid binary category targets, while CE encourages competition among sampled categories at pixels with a unique positive target. Overlapping positives and pixels without a positive target are excluded from CE.

We further supervise image-level class occurrence by
\begin{equation}
\mathcal{L}_{\mathrm{pres}}
=
\frac{1}{K}\sum_{c\in\mathcal{C}_I}
\mathcal{L}_{\mathrm{BCE}}(a_c,y_c),
\label{eq:presence_loss}
\end{equation}
where $y_c\in\{0,1\}$ indicates whether category $c$ is present in the image, as determined by the scene labels, annotated box categories, or valid semantic masks. Categories sampled from the dataset vocabulary but absent from the image annotations are treated as negative prompts with $y_c=0$.

The resulting objective has the same form for all three annotation granularities:
\begin{equation}
\mathcal{L}_{\mathrm{HS}}
=
\mathcal{L}_{\mathrm{pres}}
+
\mathcal{L}_{\mathrm{seg}}.
\label{eq:overall_loss}
\end{equation}
Each training batch contains samples from a single annotation granularity, while all batches optimize the same shared adapters under $\mathcal{L}_{\mathrm{HS}}$. Sampling rules, validity masks, and loss reductions are detailed in Sec.~\ref{sec:supp_implementation}.

Segmentation supervision is applied only to $F$, with gradients propagated through fusion into both branches; no separate branch or instance-matching losses are used. At inference, an image and category names produce independent binary masks $\mathbf{1}[F_c\geq0.5]$, without presence multiplication. Single-label benchmark evaluation uses the highest fused probability, assigning maxima below $0.5$ to background where applicable.

\subsection{Annotation-guided Online Pseudo-label Generation}
\label{sec:pseudo_labels}

We use the frozen pretrained SAM3 as the teacher, denoted by $\mathcal{T}_{\theta_0}$, to generate pseudo-labels online. The teacher processes the same augmented image and category text as the student, with its predictions computed without gradients.

\paragraph{Scene-level pseudo-labels.}
Despite the domain gap between natural and RS imagery, SAM3 retains meaningful region-level spatial structures, albeit with weak semantic activations.
Scene annotations provide reliable image-level evidence of class presence, allowing us to leverage SAM3's spatial priors to localize the corresponding regions through relative semantic responses.

For a positive category $c$, let $\overline{S}_c$ denote the teacher's semantic probability map, where $u$ indexes pixels. To account for variations in response magnitudes across images and classes, we apply robust quantile normalization:
\begin{equation}
\mathcal{N}(p)(u)
=
\operatorname{clip}
\left(
\frac{p(u)-Q_{0.02}(p)}
{\max\{Q_{0.98}(p)-Q_{0.02}(p),\epsilon\}},
0,1
\right),
\label{eq:quantile_normalization}
\end{equation}
where $Q_{\gamma}(p)$ is the $\gamma$-quantile over valid image pixels. The resulting scene pseudo-mask is
\begin{equation}
\widehat{Y}^{\mathrm{scene}}_c(u)
=
\mathbf{1}\!\left[\mathcal{N}(\overline{S}_c)(u)\geq\tau\right],
\qquad \tau=0.5.
\label{eq:scene_pseudo}
\end{equation}
This normalization mitigates the influence of extreme responses and enables pseudo-label extraction from relative semantic activations through thresholding.

For negative categories, which are randomly sampled from the remaining categories of the corresponding dataset, we assign high-confidence negative pixel targets within the union of positive scene pseudo-masks, while ignoring the remaining pixels.

\paragraph{Object-level pseudo-labels.}
Object annotations additionally provide bounding boxes, which constrain localization but do not specify precise foreground shapes. We therefore combine the teacher's Semantic and box-guided Query outputs defined in Eq.~\eqref{eq:sam3_outputs} to generate pixel-level pseudo-labels.

Let $\mathcal{B}_c$ be the retained annotated boxes of category $c$. For each box $b\in\mathcal{B}_c$, the teacher receives $(I,c,b)$ and predicts a Semantic probability map and a set of queries. We select the query that best matches both the annotated location and the category:
\begin{equation}
j_b^{*}
=
\underset{j}{\operatorname{arg\,max}}
\left[
\operatorname{IoU}(\overline{b}_{b,j},b)
+
\overline{q}_{b,j}
\right],
\label{eq:query_selection}
\end{equation}
where $\overline{b}_{b,j}$ and $\overline{q}_{b,j}$ are the teacher's predicted box and class-matching probability under prompt $b$.

We aggregate multiple boxes of the same category by pixel-wise maximum. Let $\mathcal{R}$ denote bilinear resizing to the supervision grid, and let $\overline{S}_{c,b}$ denote the semantic probability map obtained after resizing the corresponding logits:
\begin{equation}
\begin{aligned}
p^{S}_c
&=\max_{b\in\mathcal{B}_c}
\overline{S}_{c,b},\\
p^{Q}_c
&=\mathcal{R}\!\left(\max_{b\in\mathcal{B}_c}
\overline{M}_{c,b,j_b^{*}}\right).
\end{aligned}
\label{eq:object_aggregation}
\end{equation}
We normalize the aggregated Semantic response $p_c^S$ using Eq.~\eqref{eq:quantile_normalization}, while retaining the original Query mask probabilities $p_c^Q$. We combine the two responses as follows:
\begin{equation}
\widetilde{Y}^{\mathrm{object}}_c(u)
=
1-
\left[1-\mathcal{N}(p^{S}_c)(u)\right]
\left[1-p^{Q}_c(u)\right].
\label{eq:object_union}
\end{equation}
The final binary target is
\begin{equation}
\widehat{Y}^{\mathrm{object}}_c(u)
=
\mathbf{1}\!\left[
\widetilde{Y}^{\mathrm{object}}_c(u)\geq\tau
\right],
\qquad \tau=0.5.
\label{eq:object_pseudo}
\end{equation}

This construction combines relative semantic responses with geometrically guided instance masks to generate object-level pseudo-labels, enabling object annotations to contribute to the unified segmentation objective in Eq.~\eqref{eq:segmentation_loss}.


\providecommand{\HSExperimentTableStyle}{}
\definecolor{HSMeanFill}{RGB}{218,232,241}
\definecolor{HSLevelFill}{RGB}{240,246,250}
\definecolor{HSMeanText}{RGB}{30,78,108}
\definecolor{HSBestRed}{RGB}{178,43,43}
\newcolumntype{M}{>{\columncolor{HSMeanFill}\color{HSMeanText}\itshape}r}
\newcolumntype{L}{>{\columncolor{HSLevelFill}\color{HSMeanText}}r}

\providecommand{\HSresult}{\textemdash}
\providecommand{\HSyes}{\ensuremath{\checkmark}}
\providecommand{\HSno}{\textemdash}
\providecommand{\HSfigureplaceholder}[2]{%
	\fbox{\begin{minipage}[c][#1][c]{0.95\linewidth}%
			\centering\small #2\end{minipage}}}

\section{Experiments}
\label{sec:experiments}

We examine the contribution of heterogeneous supervision, compare segmentation performance, and assess cross-dataset generalization with multi-granularity inference.

\begin{figure*}[h]
    \centering
    \includegraphics[width=\textwidth]{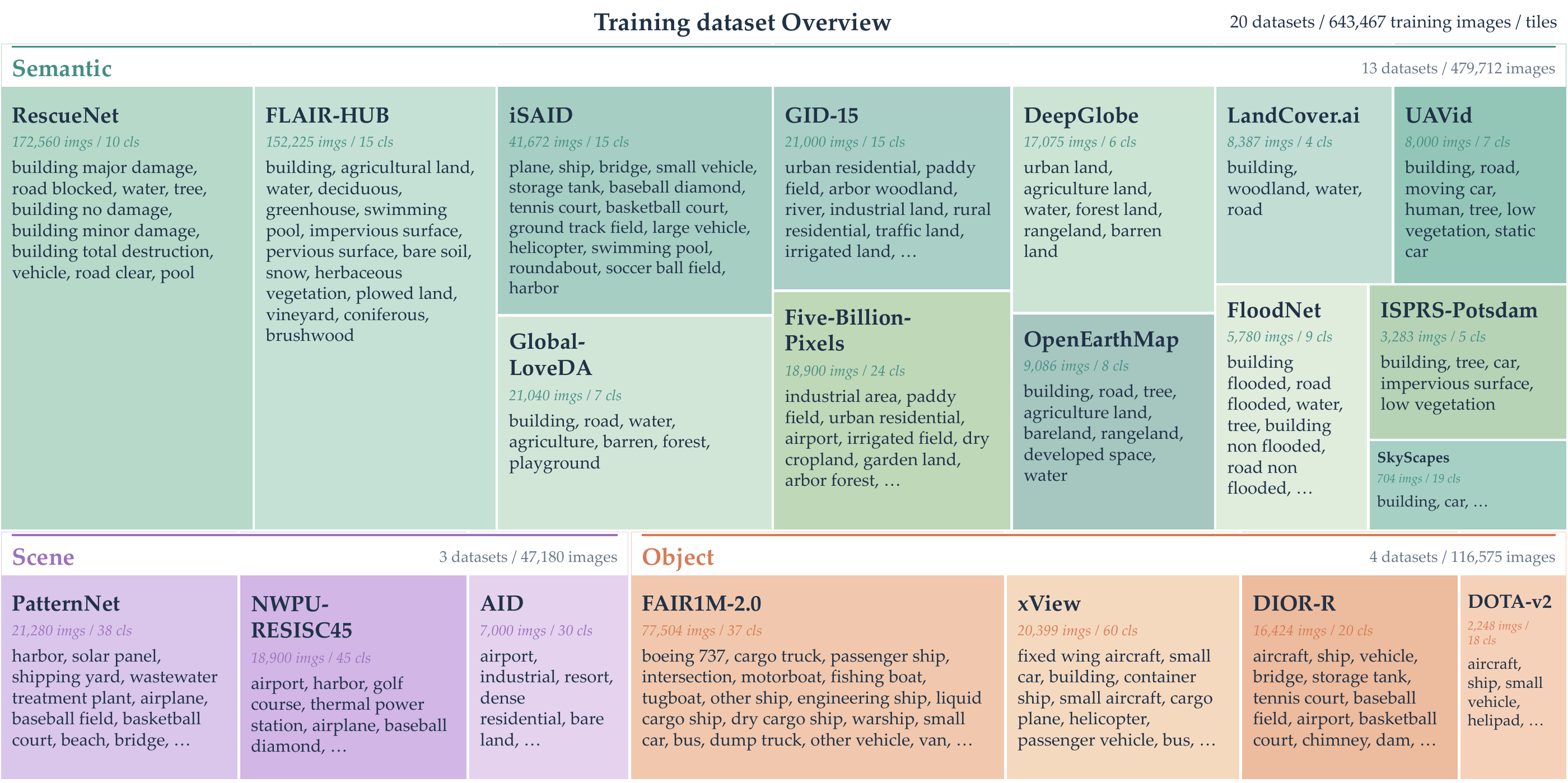}
    \vspace{-0.6cm}
    \caption{ Overview of the 20 heterogeneous training datasets used in HS-OVSeg. }
    \vspace{-0.1cm}
    \label{fig:training_dataset_distribution}
\end{figure*}

\subsection{Experimental Setup}
\label{sec:experimental_setup}

\paragraph{Datasets.}
As shown in Fig.~\ref{fig:training_dataset_distribution}, we train on 20 RS datasets comprising three scene-classification~\cite{aid_2017,nwpu_resisc45_2017,patternnet_2018}, four object-detection~\cite{dior_2020,dior_r_2022,dota_v2_2022,fair1m_2022,xview_2018}, and 13 semantic-segmentation sources~\cite{deepglobe_2018,flair_hub_2026,five_billion_pixels_2023,floodnet_2021,gid_2020,earthvl_2026,earthvqa_2024,loveda_2021,isprs_potsdam,landcover_ai_2021,openearthmap_2023,rescuenet_2023,skyscapes_2019,uavid_2020,isaid_2019}. Training dataset details are provided in Supplementary Sec.~\ref{sec:supp_experiments}. Evaluation covers held-out test sets from these sources, including 1,000 manually annotated images from the seven scene/object datasets, and five out-of-domain (OOD) semantic datasets. All methods use identical test subsets and dataset-level query vocabularies.

\paragraph{Comparison protocol.}
We use released checkpoints and method-specific preprocessing for all competing methods, with native SAM3 as the unadapted reference. Quantitative evaluation assigns each valid pixel to the highest-scoring category and reports dataset-level mIoU (\%). Scene, Object, and Pixel means weight datasets equally within each level; All 20 and OOD means weight their respective datasets equally. Source comparisons use overlapping $518\times518$ windows with stride 388. Full decoding and validity rules are given in Supplementary Sec.~\ref{sec:supp_decoding}.

\subsection{Ablation Studies}
\label{sec:ablation_experiments}

\paragraph{Quality of pseudo-labels.}
\label{sec:pseudo_label_quality}
As shown in Table~\ref{tab:pseudo_quality}, we evaluate frozen-teacher pseudo-labels (PLs) generated from coarse annotations (scene-level labels or object-level bounding boxes) against manually annotated masks using dataset-level binary mIoU. Scene- and Object-derived PLs achieve 76.24\% and 30.83\% mean mIoU, respectively. Higher Scene scores partly reflect larger regions, which are less sensitive to boundary errors. These results suggest that coarse annotations provide useful weak supervision for dense segmentation despite varying PL quality. Notably, these PLs are annotation-guided, whereas unified inference requires only images and queried category names.

\begin{table}[h]
\centering
\renewcommand{\arraystretch}{0.98}
\footnotesize
\setlength{\tabcolsep}{7pt}
\begin{tabular}{llrrM}
\toprule
Supervision & Dataset & Images & PL mIoU (\%) & Mean (\%) \\
\midrule
\multirow{3}{*}{Scene}
 & AID & 90 & 71.35 & \\
 & NWPU & 135 & 76.34 & \\
 & PatternN. & 114 & \textbf{81.02}
 & \multirow{-3}{*}{76.24} \\
\midrule
\multirow{4}{*}{Object}
 & DIOR-R & 200 & 53.05 & \\
 & DOTA-v2 & 160 & 31.38 & \\
 & FAIR1M & 201 & 19.07 & \\
 & xView & 100 & 19.83
 & \multirow{-4}{*}{30.83} \\
\bottomrule
\end{tabular}
\vspace{-0.2cm}
\caption{Coarse annotation-guided PL quality on test images. Protocol and more details: Supplementary Sec.~\ref{sec:supp_pseudo_quality}.}
\label{tab:pseudo_quality}
\vspace{2pt}
\end{table}

\begin{table}[h]
    \vspace{0.15cm}
	\centering
	\footnotesize
	\setlength{\tabcolsep}{4pt}
	\renewcommand{\arraystretch}{0.82}
	\resizebox{\columnwidth}{!}{%
		\begin{tabular}{lccc LLLM}
			\toprule
			& \multicolumn{3}{c}{Training supervision}
			 & \multicolumn{4}{c}{mIoU (\%) $\uparrow$} \\
			\cmidrule(lr){2-4}\cmidrule(lr){5-8}
			Variant & Scene & Object & Pixel & Scene & Object & Pixel & All Mean\\
			\midrule
			Native SAM3 & \HSno & \HSno & \HSno & 8.25 & 4.07 & 18.65 & 14.17 \\
			\midrule
			Scene only & \HSyes & \HSno & \HSno & 75.81 & 16.95 & 21.19 & 28.54 \\
			Object only & \HSno & \HSyes & \HSno & 23.43 & 30.09 & 19.86 & 22.44 \\
			Pixel only & \HSno & \HSno & \HSyes & 24.88 & 22.76 & 64.07 & 49.93 \\
			Scene + Object & \HSyes & \HSyes & \HSno & 74.14 & 28.38 & 25.03 & 33.07 \\
			Scene + Pixel & \HSyes & \HSno & \HSyes & \textbf{76.03} & 24.43 & \textbf{64.25} & 58.05 \\
			Object + Pixel & \HSno & \HSyes & \HSyes & 27.59 & \textbf{38.82} & 63.55 & 53.21 \\
			\midrule
			\textbf{All (HS-OVSeg)} & \HSyes & \HSyes & \HSyes & 75.12 & 38.03 & 64.13 & \textcolor{HSBestRed}{\textbf{60.56}} \\
			\bottomrule
		\end{tabular}%
	}
	\caption{Supervision complementarity with per-level exposure per epoch. More details: Supplementary Secs.~\ref{sec:supp_evaluation_protocol} and~\ref{sec:supp_target_ablation}.}
	\label{tab:supervision_complementarity}
    \vspace{-0.05cm}
\end{table}

\begin{figure}[h]
    \centering
    \includegraphics[width=0.94\columnwidth]{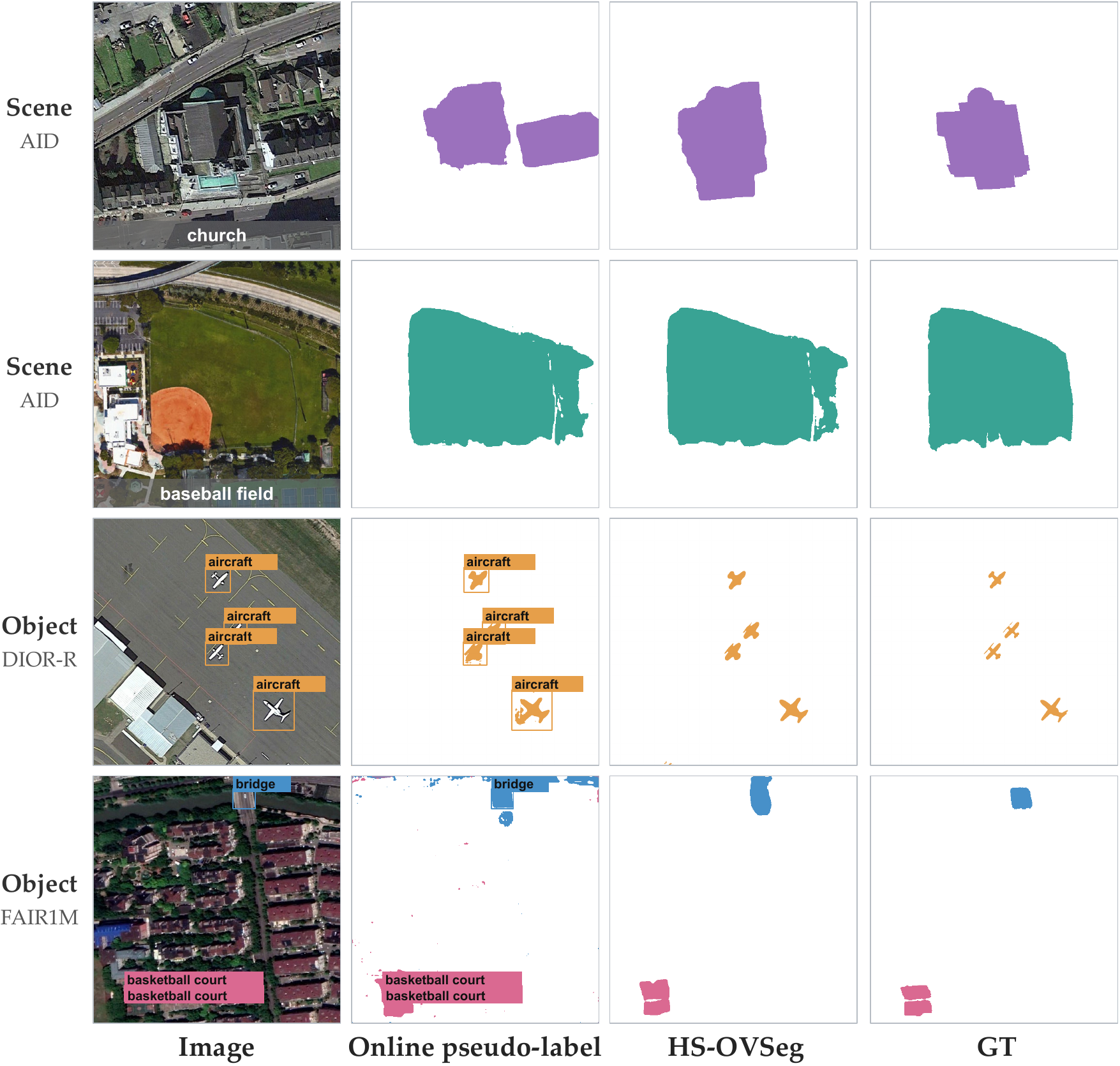}
    \caption{Teacher pseudo-labels and HS-OVSeg predictions on four illustrative examples.}
    \label{fig:pseudo_label_refinement}
    \vspace{-0.3cm}
\end{figure}

\begin{table*}[h]
\centering
\footnotesize
\renewcommand{\arraystretch}{0.82}
\setlength{\tabcolsep}{2pt}

\definecolor{HSMeanFill}{RGB}{218,232,241}
\definecolor{HSLevelFill}{RGB}{240,246,250}
\definecolor{HSMeanText}{RGB}{30,78,108}
\definecolor{HSBestRed}{RGB}{178,43,43}

\newcolumntype{U}{w{r}{8.5mm}}

\newcolumntype{D}{w{r}{9.5mm}}

\newcolumntype{E}{%
  >{\columncolor{HSLevelFill}\color{HSMeanText}}w{r}{10mm}}

\newcolumntype{O}{%
  >{\columncolor{HSMeanFill}\color{HSMeanText}\itshape}w{r}{15mm}}

\textbf{(a) Pixel-source test sets: mIoU (\%) $\uparrow$}
\par\smallskip

\begin{tabular}{@{}w{l}{28mm}*{13}{U}@{}}
\toprule
Method
& \rotatebox{30}{DeepG.-LC}
& \rotatebox{30}{FLAIR-H.}
& \rotatebox{30}{Five-B.P.}
& \rotatebox{30}{FloodNet}
& \rotatebox{30}{GID-15}
& \rotatebox{30}{G.-LoveDA}
& \rotatebox{30}{Potsdam}
& \rotatebox{30}{LandC.ai}
& \rotatebox{30}{OpenE.M.}
& \rotatebox{30}{RescueN.}
& \rotatebox{30}{SkySc.}
& \rotatebox{30}{UAVid}
& \rotatebox{30}{iSAID} \\
\midrule
Native SAM3~\cite{sam3_2026} & 12.42 & 12.13 & 6.31 & 17.61 & 5.99 & 25.09 & 33.90 & 30.56 & 27.26 & 6.43 & 20.52 & 34.16 & 10.05 \\
RemoteCLIP~\cite{remoteclip_2024} & 26.90 & 13.45 & 11.82 & 16.40 & 8.01 & 32.25 & 22.01 & 43.32 & 20.92 & 7.26 & 6.30 & 17.39 & 14.20 \\
RSKT-Seg~\cite{rskt_seg_2026} & 28.39 & 19.43 & 10.07 & 27.02 & 8.58 & 31.94 & 24.11 & 42.71 & 16.22 & 14.80 & 5.95 & 29.46 & 9.63 \\
GSNet~\cite{gsnet_2025} & 34.87 & 20.16 & 12.59 & 27.29 & 11.12 & 37.14 & 36.44 & 21.31 & 27.70 & 16.02 & 7.43 & 29.09 & 7.38 \\
LDSeg~\cite{ldseg_2026} & 32.12 & 17.07 & 15.18 & 32.63 & 15.91 & 29.36 & 34.12 & 40.09 & 30.01 & 15.66 & 13.54 & 36.92 & 10.65 \\
SegEarth-OV~\cite{segearth_ov_2025} & 30.25 & 17.27 & 14.02 & 34.14 & 16.02 & 31.14 & 36.43 & 39.60 & 31.37 & 14.27 & 14.40 & 38.90 & 13.35 \\
RSCLIP~\cite{rsclip_2026} & 31.07 & 17.46 & 14.36 & 34.72 & 15.84 & 31.36 & 36.79 & 39.67 & 31.44 & 14.98 & 14.45 & 38.95 & 13.15 \\
GeoRSCLIP~\cite{georsclip_2024} & 33.24 & 16.04 & 18.16 & 31.09 & 16.02 & 30.97 & 27.66 & 42.53 & 28.67 & 11.35 & 9.95 & 33.60 & 12.96 \\
Trident~\cite{trident_2025} & 34.52 & 19.27 & 14.89 & 36.59 & 13.96 & 29.15 & 36.47 & 44.99 & 27.73 & 18.91 & 14.13 & 40.25 & 12.24 \\
CorrCLIP~\cite{corrclip_2025} & 29.95 & 20.15 & 14.46 & 38.47 & 13.14 & 30.55 & 39.02 & 38.45 & 26.33 & 18.09 & 11.76 & 40.88 & 10.14 \\
SegEarth-OV3~\cite{segearth_ov3_2025} & 30.23 & 16.52 & 9.69 & 30.20 & 3.68 & 32.55 & 34.88 & 56.32 & 35.07 & 11.49 & 21.49 & 44.26 & 23.77 \\
CAT-Seg~\cite{cat_seg_2024} & 42.31 & 24.28 & 12.56 & 40.72 & 13.83 & 35.59 & 46.20 & 43.92 & 29.07 & 22.61 & 11.21 & 46.06 & 12.22 \\
SkySense-O~\cite{skysense_o_2025} & 41.07 & 27.34 & 15.85 & 35.94 & 6.44 & 51.85 & 38.49 & 55.26 & 34.61 & 24.50 & 11.22 & 33.30 & 32.81 \\
\midrule
\textbf{HS-OVSeg} & \textbf{71.96} & \textbf{64.61} & \textbf{65.62} & \textbf{60.17} & \textbf{48.64} & \textbf{65.12} & \textbf{81.94} & \textbf{83.54} & \textbf{66.05} & \textbf{42.54} & \textbf{44.48} & \textbf{76.58} & \textbf{62.49} \\
\bottomrule
\end{tabular}

\par\medskip
\textbf{(b) Scene- and object-source test sets and overall mean:
mIoU (\%) $\uparrow$}
\par\smallskip

\begin{tabular}{@{}w{l}{30mm}*{3}{D}|*{4}{D}|EEEO@{}}
\toprule
& \multicolumn{3}{c}{Scene sources}
& \multicolumn{4}{c}{Object sources}
& \multicolumn{4}{|c}{Means} \\
\cmidrule(lr){2-4}
\cmidrule(lr){5-8}
\cmidrule(lr){9-12}
Method
& \rotatebox{30}{AID} & \rotatebox{30}{NWPU} & \rotatebox{30}{PatternN.}
& \rotatebox{30}{DIOR-R} & \rotatebox{30}{DOTA-v2} & \rotatebox{30}{FAIR1M} & \rotatebox{30}{xView}
& P. mean & S. mean & O. mean & All mean \\
\midrule
Native SAM3~\cite{sam3_2026} & 7.25 & 7.47 & 10.04 & 8.12 & 1.61 & 1.90 & 4.64 & 18.65 & 8.25 & 4.07 & 14.17 \\
RemoteCLIP~\cite{remoteclip_2024} & 29.43 & 15.38 & 12.66 & 12.57 & 9.34 & 3.33 & 1.28 & 18.48 & 19.16 & 6.63 & 16.21 \\
RSKT-Seg~\cite{rskt_seg_2026} & 21.77 & 15.52 & 21.69 & 7.44 & 3.62 & 1.71 & 1.51 & 20.64 & 19.66 & 3.57 & 17.08 \\
GSNet~\cite{gsnet_2025} & 13.52 & 12.81 & 14.75 & 10.90 & 2.30 & 0.96 & 1.67 & 22.19 & 13.69 & 3.96 & 17.27 \\
LDSeg~\cite{ldseg_2026} & 16.14 & 13.20 & 11.99 & 8.08 & 4.23 & 2.15 & 2.43 & 24.87 & 13.77 & 4.22 & 19.07 \\
SegEarth-OV~\cite{segearth_ov_2025} & 14.39 & 12.70 & 10.99 & 9.65 & 5.47 & 2.84 & 2.25 & 25.47 & 12.69 & 5.05 & 19.47 \\
RSCLIP~\cite{rsclip_2026} & 14.81 & 13.22 & 11.30 & 9.54 & 5.74 & 2.70 & 2.27 & 25.71 & 13.11 & 5.06 & 19.69 \\
GeoRSCLIP~\cite{georsclip_2024} & 19.71 & 19.95 & 15.92 & 13.71 & 7.56 & 2.68 & 2.82 & 24.02 & 18.53 & 6.69 & 19.73 \\
Trident~\cite{trident_2025} & 18.37 & 17.73 & 13.75 & 9.54 & 6.20 & 2.26 & 1.67 & 26.39 & 16.62 & 4.92 & 20.63 \\
CorrCLIP~\cite{corrclip_2025} & 23.12 & 23.10 & 14.70 & 12.36 & 7.50 & 3.12 & 3.93 & 25.49 & 20.30 & 6.73 & 20.96 \\
SegEarth-OV3~\cite{segearth_ov3_2025} & 18.94 & 17.76 & 24.35 & 19.24 & 16.01 & 7.97 & 3.67 & 26.93 & 20.35 & 11.72 & 22.90 \\
CAT-Seg~\cite{cat_seg_2024} & 23.08 & 20.31 & 18.89 & 12.42 & 4.55 & 2.30 & 2.77 & 29.28 & 20.76 & 5.51 & 23.24 \\
SkySense-O~\cite{skysense_o_2025} & 27.46 & 35.52 & 36.25 & 51.57 & 27.64 & 10.66 & 4.39 & 31.44 & 33.08 & 23.56 & 30.11 \\
\midrule
\textbf{HS-OVSeg} & \textbf{73.90} & \textbf{71.45} & \textbf{80.00} & \textbf{63.83} & \textbf{48.86} & \textbf{21.71} & \textbf{17.71} & \textbf{64.13} & \textbf{75.12} & \textbf{38.03} & \textcolor{HSBestRed}{\textbf{60.56}} \\
\bottomrule
\end{tabular}

\vspace{-0.1cm}
\caption{Test performance on 20 training-source datasets. Additional analyses: Supplementary Secs.~\ref{sec:experimental_datasets}, \ref{sec:supp_evaluation_protocol}, and~\ref{sec:additional_analyses}.}
\label{tab:source_comparison_per_dataset}
\end{table*}
\vspace{0.1cm}

\begin{figure*}[h]
	\centering
	\includegraphics[width=\textwidth]{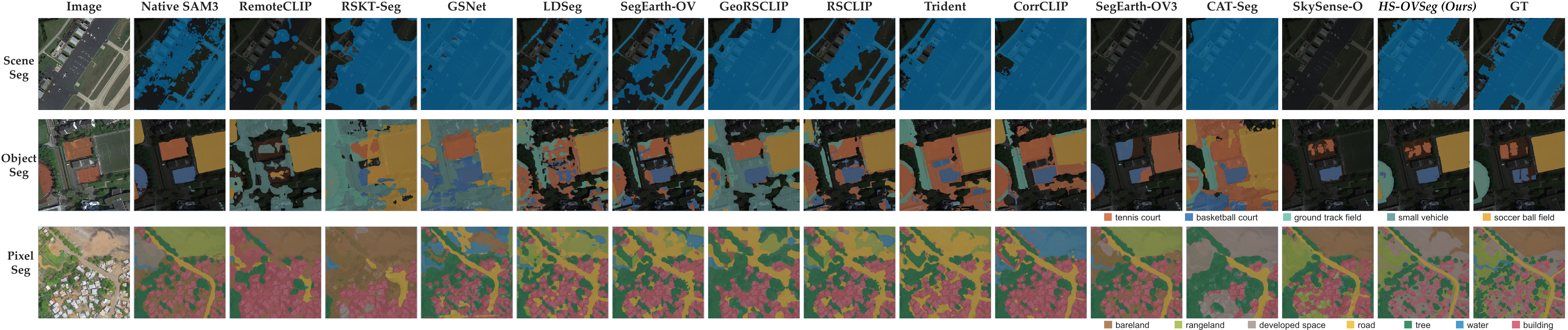}
    \vspace{-0.5cm}
    \caption{Qualitative comparisons on Scene, Object, and Pixel test samples. Additional examples are provided in Supplementary Fig. \ref{fig:supp_comparison_samples}.}
	\label{fig:qualitative_comparison}
\end{figure*}

\paragraph{Complementarity of heterogeneous supervision.}
We examine single-level and joint adaptation across all 20 test sets (Table~\ref{tab:supervision_complementarity}). Single-level adaptation improves performance on other annotation levels over native SAM3, demonstrating cross-level transfer. However, Pixel-only training achieves only 24.88\% Scene and 22.76\% Object mIoU, indicating that accurate pixel-level supervision alone cannot provide the broad semantic coverage of coarse annotations. Combining Scene and Object supervision slightly reduces their respective performance compared with single-level training (75.81\% to 74.14\% and 30.09\% to 28.38\%), potentially due to interference between noisy pseudo-labels. In contrast, incorporating Pixel supervision improves Scene mIoU from 75.81\% to 76.03\% and Object mIoU from 30.09\% to 38.82\% when paired with their respective supervision levels, suggesting that reliable pixel-level labels can benefit coarse-level adaptation. Meanwhile, incorporating both coarse annotation levels preserves Pixel performance (64.13\% vs. 64.07\% for Pixel-only). Ultimately, integrating all three levels achieves the highest overall mIoU (60.56\%), combining broad semantic coverage with accurate dense segmentation. Fig.~\ref{fig:pseudo_label_refinement} further illustrates more coherent segmentation regions than teacher-generated pseudo-labels. Additional ablations and per-class results are provided in Supplementary Sec.~\ref{sec:supp_target_ablation} and Fig.~\ref{fig:supp_class_iou}.

\subsection{Comparison with Existing Methods}
\label{sec:comparison_experiments}

We assess whether heterogeneous supervision improves dense prediction across semantic granularities by comparing HS-OVSeg with 4 general-purpose and 9 RS-oriented baselines on all 20 training-source test sets (Table~\ref{tab:source_comparison_per_dataset}). These include general-purpose dense prediction methods such as CAT-Seg, CorrCLIP, and Trident; RS vision-language backbones such as RemoteCLIP and GeoRSCLIP evaluated through dense adapters; and RS-oriented segmentation models such as GSNet, SegEarth-OV3, and SkySense-O. HS-OVSeg achieves 60.56\% overall mIoU, exceeding native SAM3 and the strongest overall baseline, SkySense-O, by 46.39 and 30.45 percentage points, respectively, with the highest score on every dataset. On Scene sources, it reaches 75.12\% mean mIoU, compared with 33.08\% for SkySense-O and 20.35\% for SegEarth-OV3. Despite RS-specific vision-language learning or SAM3-based mask fusion, these baselines remain less effective at localizing the evaluated scene concepts. Together with the single-level ablations, the Scene results support bringing scene annotations into dense supervision to extend segmentation beyond conventional land-cover vocabularies.

The Object comparison evaluates whether box-guided supervision remains effective under box-free inference. HS-OVSeg achieves 38.03\% mean mIoU, outperforming SkySense-O at 23.56\% and SegEarth-OV3 at 11.72\%. On DIOR-R and DOTA-v2, it improves over SkySense-O from 51.57\% to 63.83\% and from 27.64\% to 48.86\%, respectively, highlighting the benefits of learning from object annotations without requiring boxes at inference. However, FAIR1M and xView remain challenging, with mIoUs of 21.71\% and 17.71\%, respectively, due to fine-grained categories and small objects. On Pixel sources, HS-OVSeg achieves 64.13\% mean mIoU, surpassing SkySense-O at 31.44\% and CAT-Seg at 29.28\%, with performance across land-cover datasets and iSAID. Together with the supervision-complementarity results, these findings demonstrate that heterogeneous supervision improves object-level segmentation while preserving Pixel performance.

Fig.~\ref{fig:qualitative_comparison} illustrates improved segmentation across all three groups. Qualitative results use GT-present prompts, while quantitative evaluation uses the full dataset vocabulary.

\subsection{Out-of-Domain Evaluation}
\label{sec:held_out_generalization}

To assess generalization beyond the adaptation sources, we evaluate the same HS-OVSeg checkpoint on five datasets without target-dataset fine-tuning, as reported in Table~\ref{tab:ood_comparison}.
HS-OVSeg achieves the highest mean mIoU of 43.33\%, outperforming native SAM3 and SegEarth-OV3 by 13.91 and 6.22 percentage points, respectively. HS-OVSeg leads on DLRSD, WHDLD, and VDD, whereas SegEarth-OV3 leads on AeroScapes and SkySense-O on DynamicEarthNet. The results support transfer across held-out domains while showing that the benefits depend on the target dataset.

\begin{table}[ht]
	\centering
	\footnotesize
	\setlength{\tabcolsep}{3pt}
	\renewcommand{\arraystretch}{0.82}
	\resizebox{\columnwidth}{!}{%
		\begin{tabular}{lrrrrr|M}
			\toprule
			& \multicolumn{6}{c}{mIoU (\%) $\uparrow$} \\
			\cmidrule(lr){2-7}
			Method & DLRSD & AeroSc. & WHDLD & VDD & Dyn.E. & Mean \\
			\midrule
RSKT-Seg~\cite{rskt_seg_2026} & 27.78 & 17.13 & 20.19 & 30.57 & 18.05 & 22.74 \\
RemoteCLIP~\cite{remoteclip_2024} & 25.69 & 12.48 & 31.47 & 28.16 & 17.36 & 23.03 \\
GSNet~\cite{gsnet_2025} & 22.80 & 15.70 & 29.26 & 31.51 & 18.71 & 23.60 \\
GeoRSCLIP~\cite{georsclip_2024} & 26.54 & 28.17 & 25.80 & 27.62 & 26.51 & 26.93 \\
Native SAM3~\cite{sam3_2026} & 22.34 & 37.81 & 40.91 & 29.81 & 16.24 & 29.42 \\
LDSeg~\cite{ldseg_2026} & 25.86 & 37.96 & 24.31 & 33.45 & 28.78 & 30.07 \\
CorrCLIP~\cite{corrclip_2025} & 31.01 & 42.75 & 28.43 & 38.03 & 14.87 & 31.02 \\
Trident~\cite{trident_2025} & 28.16 & 44.39 & 21.91 & 36.37 & 28.29 & 31.82 \\
SegEarth-OV~\cite{segearth_ov_2025} & 28.61 & 40.64 & 29.01 & 34.62 & 29.75 & 32.53 \\
RSCLIP~\cite{rsclip_2026} & 28.97 & 41.23 & 28.69 & 35.41 & 29.81 & 32.82 \\
CAT-Seg~\cite{cat_seg_2024} & 33.05 & 47.39 & 27.57 & 37.61 & 18.39 & 32.80 \\
SkySense-O~\cite{skysense_o_2025} & 40.89 & 23.13 & 30.68 & 39.90 & \textbf{36.50} & 34.22 \\
SegEarth-OV3~\cite{segearth_ov3_2025} & 36.74 & \textbf{52.22} & 42.87 & 44.45 & 9.23 & 37.10 \\
\midrule
\textbf{HS-OVSeg} & \textbf{45.39} & 45.54 & \textbf{43.96} & \textbf{52.65} & 29.10 & \textcolor{HSBestRed}{\textbf{43.33}} \\
			\bottomrule
		\end{tabular}%
	}
\vspace{-0.1cm}
\caption{OOD performance comparison on 5 held-out datasets. Dataset details: Supplementary Sec.~\ref{sec:supp_ood_evaluation}.}
\vspace{0.05cm}
\label{tab:ood_comparison}
\end{table}

As illustrated in Fig.~\ref{fig:multi_granularity_prediction}, HS-OVSeg generates multi-granularity segmentation maps from the same OOD NAIP aerial image acquired over San Diego, USA, which illustrates unified text-driven segmentation.

\section{Conclusion}
\label{sec:conclusion}
We presented HS-OVSeg, a unified multi-granularity framework for RS open-vocabulary semantic segmentation integrating scene-, object-, and pixel-level supervision. By generating pixel-level pseudo-labels from coarse annotations using SAM3 and optimizing shared low-rank adapters, HS-OVSeg enables learning across annotation granularities. Experiments on 20 training-source datasets demonstrate substantial improvements over existing methods, particularly for scene- and object-level segmentation, while maintaining strong pixel-level performance. Evaluation on five held-out datasets further demonstrates cross-dataset generalization. These results highlight the potential of heterogeneous supervision for multi-granularity RS segmentation.

Despite these advances, segmenting small objects and distinguishing fine-grained categories remain challenging, particularly on FAIR1M and xView, motivating further research on fine-grained recognition.

\section*{LLM Usage Statement}
ChatGPT was used to assist with English language editing, grammatical corrections, and improving the clarity and organization of the manuscript. It was also used to discuss and refine the presentation of methodological descriptions. All scientific ideas, technical implementations, experimental results, and conclusions were developed, verified, and validated by the authors.

{
    \small
    \bibliographystyle{ieeenat_fullname}
    \bibliography{main}

@article{ldseg_2026,
  title={Exploiting Foundation Models for Costless Open-Vocabulary Semantic Segmentation of Optical Remote Sensing Images},
  author={Ye, Bowei and Shao, Tao and Su, Fei and Zhao, Zhicheng},
  journal={IEEE Transactions on Geoscience and Remote Sensing},
  volume={64},
  pages={1--14},
  year={2026},
  publisher={IEEE}
}

@inproceedings{sam3_2026,
  title={Sam 3: Segment anything with concepts},
  author={Carion, Nicolas and Gustafson, Laura and Hu, Yuan-Ting and Debnath, Shoubhik and Hu, Ronghang and Suris Coll-Vinent, Didac and Ryali, Chaitanya and Alwala, Kalyan Vasudev and Khedr, Haitham and Huang, Andrew and others},
  booktitle={International conference on learning representations},
  volume={2026},
  pages={138846--138923},
  year={2026}
}

@inproceedings{proxyclip_2024,
  title={Proxyclip: Proxy attention improves clip for open-vocabulary segmentation},
  author={Lan, Mengcheng and Chen, Chaofeng and Ke, Yiping and Wang, Xinjiang and Feng, Litong and Zhang, Wayne},
  booktitle={European Conference on Computer Vision},
  pages={70--88},
  year={2024},
  organization={Springer}
}

@inproceedings{trident_2025,
  title={Harnessing vision foundation models for high-performance, training-free open vocabulary segmentation},
  author={Shi, Yuheng and Dong, Minjing and Xu, Chang},
  booktitle={2025 IEEE/CVF International Conference on Computer Vision (ICCV)},
  pages={23487--23497},
  year={2025},
  organization={IEEE}
}

@inproceedings{corrclip_2025,
  title={Corrclip: Reconstructing patch correlations in clip for open-vocabulary semantic segmentation},
  author={Zhang, Dengke and Liu, Fagui and Tang, Quan},
  booktitle={2025 IEEE/CVF International Conference on Computer Vision (ICCV)},
  pages={24677--24687},
  year={2025},
  organization={IEEE}
}

@inproceedings{maskclip_2022,
  title={Extract free dense labels from clip},
  author={Zhou, Chong and Loy, Chen Change and Dai, Bo},
  booktitle={European conference on computer vision},
  pages={696--712},
  year={2022},
  organization={Springer}
}

@inproceedings{gem_2024,
  title={Grounding everything: Emerging localization properties in vision-language transformers},
  author={Bousselham, Walid and Petersen, Felix and Ferrari, Vittorio and Kuehne, Hilde},
  booktitle={2024 IEEE/CVF Conference on Computer Vision and Pattern Recognition (CVPR)},
  pages={3828--3837},
  year={2024},
  organization={IEEE}
}

@inproceedings{sclip_2024,
  title={Sclip: Rethinking self-attention for dense vision-language inference},
  author={Wang, Feng and Mei, Jieru and Yuille, Alan},
  booktitle={European conference on computer vision},
  pages={315--332},
  year={2024},
  organization={Springer}
}

@inproceedings{lan2024clearclip,
  title={Clearclip: Decomposing clip representations for dense vision-language inference},
  author={Lan, Mengcheng and Chen, Chaofeng and Ke, Yiping and Wang, Xinjiang and Feng, Litong and Zhang, Wayne},
  booktitle={European Conference on Computer Vision},
  pages={143--160},
  year={2024},
  organization={Springer}
}

@inproceedings{naclip_2025,
  title={Pay attention to your neighbours: Training-free open-vocabulary semantic segmentation},
  author={Hajimiri, Sina and Ben Ayed, Ismail and Dolz, Jose},
  booktitle={Proceedings of the winter conference on applications of computer vision},
  pages={5061--5071},
  year={2025}
}

@article{segearth_ov3_2025,
  title={SegEarth-OV3: Exploring SAM 3 for open-vocabulary semantic segmentation in remote sensing images},
  author={Li, Kaiyu and Zhang, Shengqi and Wang, Yujie and Deng, Yupeng and Wang, Zhi and Meng, Deyu and Cao, Xiangyong},
  journal={arXiv preprint arXiv:2512.08730},
  year={2025}
}

@inproceedings{segearth_ov_2025,
  title={Segearth-ov: Towards training-free open-vocabulary segmentation for remote sensing images},
  author={Li, Kaiyu and Liu, Ruixun and Cao, Xiangyong and Bai, Xueru and Zhou, Feng and Meng, Deyu and Wang, Zhi},
  booktitle={2025 IEEE/CVF Conference on Computer Vision and Pattern Recognition (CVPR)},
  pages={10545--10556},
  year={2025},
  organization={IEEE}
}

@inproceedings{cat_seg_2024,
  title={Cat-seg: Cost aggregation for open-vocabulary semantic segmentation},
  author={Cho, Seokju and Shin, Heeseong and Hong, Sunghwan and Arnab, Anurag and Seo, Paul Hongsuck and Kim, Seungryong},
  booktitle={2024 IEEE/CVF Conference on Computer Vision and Pattern Recognition (CVPR)},
  pages={4113--4123},
  year={2024},
  organization={IEEE}
}

@inproceedings{ebseg_2024,
  title={Open-vocabulary semantic segmentation with image embedding balancing},
  author={Shan, Xiangheng and Wu, Dongyue and Zhu, Guilin and Shao, Yuanjie and Sang, Nong and Gao, Changxin},
  booktitle={2024 IEEE/CVF Conference on Computer Vision and Pattern Recognition (CVPR)},
  pages={28412--28421},
  year={2024},
  organization={IEEE}
}

@article{remoteclip_2024,
  title={Remoteclip: A vision language foundation model for remote sensing},
  author={Liu, Fan and Chen, Delong and Guan, Zhangqingyun and Zhou, Xiaocong and Zhu, Jiale and Ye, Qiaolin and Fu, Liyong and Zhou, Jun},
  journal={IEEE Transactions on Geoscience and Remote Sensing},
  volume={62},
  pages={1--16},
  year={2024},
  publisher={IEEE}
}

@article{georsclip_2024,
  title={RS5M and GeoRSCLIP: A large-scale vision-language dataset and a large vision-language model for remote sensing},
  author={Zhang, Zilun and Zhao, Tiancheng and Guo, Yulong and Yin, Jianwei},
  journal={IEEE Transactions on Geoscience and Remote Sensing},
  volume={62},
  pages={1--23},
  year={2024},
  publisher={IEEE}
}

@article{rsclip_2026,
  title={RSCLIP for Training-Free Open-Vocabulary Remote Sensing Image Semantic Segmentation},
  author={Wang, Shanwen and Sun, Xin and Han, Jungong and Zhu, Xiao Xiang},
  journal={IEEE Journal of Selected Topics in Applied Earth Observations and Remote Sensing},
  year={2026},
  publisher={IEEE}
}

@inproceedings{skysense_o_2025,
  title={Skysense-o: Towards open-world remote sensing interpretation with vision-centric visual-language modeling},
  author={Zhu, Qi and Lao, Jiangwei and Ji, Deyi and Luo, Junwei and Wu, Kang and Zhang, Yingying and Ru, Lixiang and Wang, Jian and Chen, Jingdong and Yang, Ming and others},
  booktitle={2025 IEEE/CVF Conference on Computer Vision and Pattern Recognition (CVPR)},
  pages={14733--14744},
  year={2025},
  organization={IEEE}
}

@inproceedings{openseed_2023,
  title={A simple framework for open-vocabulary segmentation and detection},
  author={Zhang, Hao and Li, Feng and Zou, Xueyan and Liu, Shilong and Li, Chunyuan and Yang, Jianwei and Zhang, Lei},
  booktitle={2023 IEEE/CVF International Conference on Computer Vision (ICCV)},
  pages={1020--1031},
  year={2023},
  organization={IEEE}
}

@inproceedings{openseg_2022,
  title={Scaling open-vocabulary image segmentation with image-level labels},
  author={Ghiasi, Golnaz and Gu, Xiuye and Cui, Yin and Lin, Tsung-Yi},
  booktitle={European conference on computer vision (ECCV)},
  pages={540--557},
  year={2022},
  organization={Springer}
}

@inproceedings{lora_2022,
  title={{LoRA}: Low-Rank Adaptation of Large Language Models},
  author={Hu, Edward J. and Shen, Yelong and Wallis, Phillip and Allen-Zhu, Zeyuan and Li, Yuanzhi and Wang, Shean and Wang, Lu and Chen, Weizhu},
  booktitle={International Conference on Learning Representations (ICLR)},
  year={2022},
  url={https://openreview.net/forum?id=nZeVKeeFYf9}
}

@inproceedings{rskt_seg_2026,
  title={Exploring efficient open-vocabulary segmentation in the remote sensing},
  author={Li, Bingyu and Dong, Haocheng and Zhang, Da and Zhao, Zhiyuan and Sun, Hao and Gao, Junyu},
  booktitle={Proceedings of the AAAI Conference on Artificial Intelligence},
  volume={40},
  number={8},
  pages={5982--5991},
  year={2026}
}

@inproceedings{reattnclip_2026,
  title={ReAttnCLIP: Training-Free Open-Vocabulary Remote Sensing Image Segmentation via Re-defined Attention in CLIP},
  author={Niu, Xin and Zhao, Manqi and Jiang, Dongsheng and Wu, Yingying and Su, Bing},
  booktitle={Proceedings of the IEEE/CVF Conference on Computer Vision and Pattern Recognition},
  pages={24980--24989},
  year={2026}
}

@inproceedings{clip_2021,
  title={Learning transferable visual models from natural language supervision},
  author={Radford, Alec and Kim, Jong Wook and Hallacy, Chris and Ramesh, Aditya and Goh, Gabriel and Agarwal, Sandhini and Sastry, Girish and Askell, Amanda and Mishkin, Pamela and Clark, Jack and others},
  booktitle={International conference on machine learning},
  pages={8748--8763},
  year={2021},
  organization={PMLR}
}

@inproceedings{ovseg_2023,
  title={Open-vocabulary semantic segmentation with mask-adapted clip},
  author={Liang, Feng and Wu, Bichen and Dai, Xiaoliang and Li, Kunpeng and Zhao, Yinan and Zhang, Hang and Zhang, Peizhao and Vajda, Peter and Marculescu, Diana},
  booktitle={2023 IEEE/CVF Conference on Computer Vision and Pattern Recognition (CVPR)},
  pages={7061--7070},
  year={2023},
  organization={IEEE}
}

@article{san_2023,
  title={SAN: Side adapter network for open-vocabulary semantic segmentation},
  author={Xu, Mengde and Zhang, Zheng and Wei, Fangyun and Hu, Han and Bai, Xiang},
  journal={IEEE Transactions on Pattern Analysis and Machine Intelligence},
  volume={45},
  number={12},
  pages={15546--15561},
  year={2023},
  publisher={IEEE}
}

@inproceedings{groupvit_2022,
  title={Groupvit: Semantic segmentation emerges from text supervision},
  author={Xu, Jiarui and De Mello, Shalini and Liu, Sifei and Byeon, Wonmin and Breuel, Thomas and Kautz, Jan and Wang, Xiaolong},
  booktitle={2022 IEEE/CVF Conference on Computer Vision and Pattern Recognition (CVPR)},
  pages={18113--18123},
  year={2022},
  organization={IEEE}
}

@inproceedings{xdecoder_2023,
  title={Generalized decoding for pixel, image, and language},
  author={Zou, Xueyan and Dou, Zi-Yi and Yang, Jianwei and Gan, Zhe and Li, Linjie and Li, Chunyuan and Dai, Xiyang and Behl, Harkirat and Wang, Jianfeng and Yuan, Lu and others},
  booktitle={2023 IEEE/CVF Conference on Computer Vision and Pattern Recognition (CVPR)},
  pages={15116--15127},
  year={2023},
  organization={IEEE}
}

@article{seem_2023,
  title={Segment everything everywhere all at once},
  author={Zou, Xueyan and Yang, Jianwei and Zhang, Hao and Li, Feng and Li, Linjie and Wang, Jianfeng and Wang, Lijuan and Gao, Jianfeng and Lee, Yong Jae},
  journal={Advances in neural information processing systems},
  volume={36},
  pages={19769--19782},
  year={2023}
}

@inproceedings{omgseg_2024,
  title={Omg-seg: Is one model good enough for all segmentation?},
  author={Li, Xiangtai and Yuan, Haobo and Li, Wei and Ding, Henghui and Wu, Size and Zhang, Wenwei and Li, Yining and Chen, Kai and Loy, Chen Change},
  booktitle={2024 IEEE/CVF Conference on Computer Vision and Pattern Recognition (CVPR)},
  pages={27948--27959},
  year={2024},
  organization={IEEE}
}

@inproceedings{querymeldnet_2025,
  title={Scaling up image segmentation across data and tasks},
  author={Wang, Pei and Cai, Zhaowei and Yang, Hao and Swaminathan, Ashwin and Manmatha, R and Soatto, Stefano},
  booktitle={2025 IEEE/CVF Conference on Computer Vision and Pattern Recognition (CVPR)},
  pages={4573--4583},
  year={2025},
  organization={IEEE}
}

@article{dior_r_2022,
  title={Anchor-free oriented proposal generator for object detection},
  author={Cheng, Gong and Wang, Jiabao and Li, Ke and Xie, Xingxing and Lang, Chunbo and Yao, Yanqing and Han, Junwei},
  journal={IEEE Transactions on Geoscience and Remote Sensing},
  volume={60},
  pages={1--11},
  year={2022},
  publisher={IEEE}
}

@article{dota_v2_2022,
  title={Object detection in aerial images: A large-scale benchmark and challenges},
  author={Ding, Jian and Xue, Nan and Xia, Gui-Song and Bai, Xiang and Yang, Wen and Yang, Michael Ying and Belongie, Serge and Luo, Jiebo and Datcu, Mihai and Pelillo, Marcello and others},
  journal={IEEE transactions on pattern analysis and machine intelligence},
  volume={44},
  number={11},
  pages={7778--7796},
  year={2021},
  publisher={IEEE}
}

@article{flair_hub_2026,
  title={FLAIR-HUB: Large-scale multimodal dataset for land cover and crop mapping},
  author={Garioud, Anatol and Giordano, S{\'e}bastien and David, Nicolas and Gonthier, Nicolas},
  journal={ISPRS Journal of Photogrammetry and Remote Sensing},
  volume={237},
  pages={271--300},
  year={2026},
  publisher={Elsevier}
}

@article{floodnet_2021,
  title={Floodnet: A high resolution aerial imagery dataset for post flood scene understanding},
  author={Rahnemoonfar, Maryam and Chowdhury, Tashnim and Sarkar, Argho and Varshney, Debvrat and Yari, Masoud and Murphy, Robin Roberson},
  journal={IEEE Access},
  volume={9},
  pages={89644--89654},
  year={2021},
  publisher={IEEE}
}

@article{uavid_2020,
  title={UAVid: A semantic segmentation dataset for UAV imagery},
  author={Lyu, Ye and Vosselman, George and Xia, Gui-Song and Yilmaz, Alper and Yang, Michael Ying},
  journal={ISPRS journal of photogrammetry and remote sensing},
  volume={165},
  pages={108--119},
  year={2020},
  publisher={Elsevier}
}

@article{dlrsd_2018,
  title={Performance evaluation of single-label and multi-label remote sensing image retrieval using a dense labeling dataset},
  author={Shao, Zhenfeng and Yang, Ke and Zhou, Weixun},
  journal={Remote Sensing},
  volume={10},
  number={6},
  pages={964},
  year={2018},
  publisher={MDPI}
}

@article{whdld_2020,
  title={Multilabel remote sensing image retrieval based on fully convolutional network},
  author={Shao, Zhenfeng and Zhou, Weixun and Deng, Xueqing and Zhang, Maoding and Cheng, Qimin},
  journal={IEEE Journal of Selected Topics in Applied Earth Observations and Remote Sensing},
  volume={13},
  pages={318--328},
  year={2020},
  publisher={IEEE}
}

@article{vdd_2025,
  title={Vdd: Varied drone dataset for semantic segmentation},
  author={Cai, Wenxiao and Jin, Ke and Hou, Jinyan and Guo, Cong and Wu, Letian and Yang, Wankou},
  journal={Journal of Visual Communication and Image Representation},
  volume={109},
  pages={104429},
  year={2025},
  publisher={Elsevier}
}

@article{five_billion_pixels_2023,
  title={Enabling country-scale land cover mapping with meter-resolution satellite imagery},
  author={Tong, Xin-Yi and Xia, Gui-Song and Zhu, Xiao Xiang},
  journal={ISPRS Journal of Photogrammetry and Remote Sensing},
  volume={196},
  pages={178--196},
  year={2023},
  publisher={Elsevier}
}

@article{nwpu_resisc45_2017,
  title={Remote sensing image scene classification: Benchmark and state of the art},
  author={Cheng, Gong and Han, Junwei and Lu, Xiaoqiang},
  journal={Proceedings of the IEEE},
  volume={105},
  number={10},
  pages={1865--1883},
  year={2017},
  publisher={IEEE}
}

@inproceedings{deepglobe_2018,
  title={Deepglobe 2018: A challenge to parse the earth through satellite images},
  author={Demir, Ilke and Koperski, Krzysztof and Lindenbaum, David and Pang, Guan and Huang, Jing and Basu, Saikat and Hughes, Forest and Tuia, Devis and Raskar, Ramesh},
  booktitle={2018 IEEE/CVF Conference on Computer Vision and Pattern Recognition Workshops (CVPRW)},
  pages={172--181},
  year={2018},
  organization={IEEE}
}

@article{rescuenet_2023,
  title={RescueNet: A high resolution UAV semantic segmentation dataset for natural disaster damage assessment},
  author={Rahnemoonfar, Maryam and Chowdhury, Tashnim and Murphy, Robin},
  journal={Scientific data},
  volume={10},
  number={1},
  pages={913},
  year={2023},
  publisher={Nature Publishing Group UK, London}
}

@article{gid_2020,
  title={Land-cover classification with high-resolution remote sensing images using transferable deep models},
  author={Tong, Xin-Yi and Xia, Gui-Song and Lu, Qikai and Shen, Huanfeng and Li, Shengyang and You, Shucheng and Zhang, Liangpei},
  journal={Remote Sensing of Environment},
  volume={237},
  pages={111322},
  year={2020},
  publisher={Elsevier}
}

@inproceedings{skyscapes_2019,
  title={Skyscapes fine-grained semantic understanding of aerial scenes},
  author={Azimi, Seyed Majid and Henry, Corentin and Sommer, Lars and Schumann, Arne and Vig, Eleonora},
  booktitle={2019 IEEE/CVF International Conference on Computer Vision (ICCV)},
  pages={7392--7402},
  year={2019},
  organization={IEEE}
}

@inproceedings{aeroscapes_2018,
  title={Ensemble knowledge transfer for semantic segmentation},
  author={Nigam, Ishan and Huang, Chen and Ramanan, Deva},
  booktitle={2018 IEEE Winter Conference on Applications of Computer Vision (WACV)},
  pages={1499--1508},
  year={2018},
  organization={IEEE}
}

@inproceedings{earthvqa_2024,
  title={Earthvqa: Towards queryable earth via relational reasoning-based remote sensing visual question answering},
  author={Wang, Junjue and Zheng, Zhuo and Chen, Zihang and Ma, Ailong and Zhong, Yanfei},
  booktitle={Proceedings of the AAAI conference on artificial intelligence},
  volume={38},
  number={6},
  pages={5481--5489},
  year={2024}
}

@inproceedings{landcover_ai_2021,
  title={LandCover. ai: Dataset for automatic mapping of buildings, woodlands, water and roads from aerial imagery},
  author={Boguszewski, Adrian and Batorski, Dominik and Ziemba-Jankowska, Natalia and Dziedzic, Tomasz and Zambrzycka, Anna},
  booktitle={2021 IEEE/CVF Conference on Computer Vision and Pattern Recognition Workshops (CVPRW)},
  pages={1102--1110},
  year={2021},
  organization={IEEE}
}

@inproceedings{openearthmap_2023,
  title={Openearthmap: A benchmark dataset for global high-resolution land cover mapping},
  author={Xia, Junshi and Yokoya, Naoto and Adriano, Bruno and Broni-Bediako, Clifford},
  booktitle={2023 IEEE/CVF Winter Conference on Applications of Computer Vision (WACV)},
  pages={6243--6253},
  year={2023},
  organization={IEEE}
}

@article{dior_2020,
  title={Object detection in optical remote sensing images: A survey and a new benchmark},
  author={Li, Ke and Wan, Gang and Cheng, Gong and Meng, Liqiu and Han, Junwei},
  journal={ISPRS journal of photogrammetry and remote sensing},
  volume={159},
  pages={296--307},
  year={2020},
  publisher={Elsevier}
}

@article{aid_2017,
  title={AID: A benchmark data set for performance evaluation of aerial scene classification},
  author={Xia, Gui-Song and Hu, Jingwen and Hu, Fan and Shi, Baoguang and Bai, Xiang and Zhong, Yanfei and Zhang, Liangpei and Lu, Xiaoqiang},
  journal={IEEE transactions on geoscience and remote sensing},
  volume={55},
  number={7},
  pages={3965--3981},
  year={2017},
  publisher={IEEE}
}

@article{patternnet_2018,
  title={PatternNet: A benchmark dataset for performance evaluation of remote sensing image retrieval},
  author={Zhou, Weixun and Newsam, Shawn and Li, Congmin and Shao, Zhenfeng},
  journal={ISPRS journal of photogrammetry and remote sensing},
  volume={145},
  pages={197--209},
  year={2018},
  publisher={Elsevier}
}

@article{fair1m_2022,
  title={FAIR1M: A benchmark dataset for fine-grained object recognition in high-resolution remote sensing imagery},
  author={Sun, Xian and Wang, Peijin and Yan, Zhiyuan and Xu, Feng and Wang, Ruiping and Diao, Wenhui and Chen, Jin and Li, Jihao and Feng, Yingchao and Xu, Tao and others},
  journal={ISPRS Journal of Photogrammetry and Remote Sensing},
  volume={184},
  pages={116--130},
  year={2022},
  publisher={Elsevier}
}

@article{xview_2018,
  title={xview: Objects in context in overhead imagery},
  author={Lam, Darius and Kuzma, Richard and McGee, Kevin and Dooley, Samuel and Laielli, Michael and Klaric, Matthew and Bulatov, Yaroslav and McCord, Brendan},
  journal={arXiv preprint arXiv:1802.07856},
  year={2018}
}

@article{earthvl_2026,
  title={Earthvl: A progressive earth vision-language understanding and generation framework},
  author={Wang, Junjue and Zhong, Yanfei and Chen, Zihang and Zheng, Zhuo and Ma, Ailong and Zhang, Liangpei},
  journal={arXiv preprint arXiv:2601.02783},
  year={2026}
}

@article{loveda_2021,
  title={LoveDA: A remote sensing land-cover dataset for domain adaptive semantic segmentation},
  author={Wang, Junjue and Zheng, Zhuo and Ma, Ailong and Lu, Xiaoyan and Zhong, Yanfei},
  journal={arXiv preprint arXiv:2110.08733},
  year={2021}
}

@article{isprs_potsdam,
  title={The ISPRS benchmark on urban object classification and 3D building reconstruction},
  author={Rottensteiner, Franz and Sohn, Gunho and Jung, Jaewook and Gerke, Markus and Baillard, Caroline and Benitez, Sebastien and Breitkopf, Uwe},
  journal={ISPRS Annals of the Photogrammetry, Remote Sensing and Spatial Information Sciences},
  volume={1},
  pages={293--298},
  year={2012},
  publisher={Copernicus GmbH}
}

@article{isaid_2019,
  title={isaid: A large-scale dataset for instance segmentation in aerial images},
  author={Zamir, Syed Waqas and Arora, Aditya and Gupta, Akshita and Khan, Salman and Sun, Guolei and Khan, Fahad Shahbaz and Zhu, Fan and Shao, Ling and Xia, Gui-Song and Bai, Xiang},
  journal={arXiv preprint arXiv:1905.12886},
  year={2019}
}

@inproceedings{gsnet_2025,
  title={Towards open-vocabulary remote sensing image semantic segmentation},
  author={Ye, Chengyang and Zhuge, Yunzhi and Zhang, Pingping},
  booktitle={Proceedings of the AAAI Conference on Artificial Intelligence},
  volume={39},
  number={9},
  pages={9436--9444},
  year={2025}
}
}

\clearpage
\maketitlesupplementary

\ifdefined\HSExperimentTableStyle\else
\providecommand{\HSExperimentTableStyle}{}
\definecolor{HSMeanFill}{RGB}{218,232,241}
\definecolor{HSLevelFill}{RGB}{240,246,250}
\definecolor{HSMeanText}{RGB}{30,78,108}
\definecolor{HSBestRed}{RGB}{178,43,43}
\newcolumntype{M}{>{\columncolor{HSMeanFill}\color{HSMeanText}\itshape}r}
\newcolumntype{L}{>{\columncolor{HSLevelFill}\color{HSMeanText}}r}

\providecommand{\HSresult}{\textemdash}
\providecommand{\HSyes}{\ensuremath{\checkmark}}
\providecommand{\HSno}{\textemdash}
\providecommand{\HSfigureplaceholder}[2]{%
  \fbox{\begin{minipage}[c][#1][c]{0.95\linewidth}%
  \centering\small #2\end{minipage}}}

\fi

\section{Experimental Setup}
\label{sec:supp_experiments}

\subsection{Training Sources}
\label{sec:experimental_datasets}

\paragraph{Heterogeneous training sources.}
We use 20 RS datasets: three scene-classification, four object-detection, and 13 semantic-segmentation sources. Figure~\ref{fig:training_dataset_distribution} summarizes their sample counts, categories, and supervision granularities. The prepared splits contain 47,180 scene-labeled, 116,575 box-labeled, and 479,712 pixel-labeled samples, totaling 643,467 images or tiles. These counts represent prepared training entries, not unique full-resolution images or repeated sampling draws. Scene and detection sources provide only image-level and box-level annotations for adaptation; manually annotated semantic masks are reserved for testing. The scene sources are AID~\cite{aid_2017}, NWPU-RESISC45~\cite{nwpu_resisc45_2017}, and PatternNet~\cite{patternnet_2018}. The detection sources are DIOR-R~\cite{dior_2020,dior_r_2022}, DOTA-v2~\cite{dota_v2_2022}, FAIR1M-2.0~\cite{fair1m_2022}, and xView~\cite{xview_2018}. The pixel-labeled sources are DeepGlobe-LandCover~\cite{deepglobe_2018}, FLAIR-HUB~\cite{flair_hub_2026}, Five-Billion-Pixels~\cite{five_billion_pixels_2023}, FloodNet~\cite{floodnet_2021}, GID-15~\cite{gid_2020}, Global-LoveDA~\cite{earthvl_2026,earthvqa_2024,loveda_2021}, ISPRS-Potsdam~\cite{isprs_potsdam}, LandCover.ai~\cite{landcover_ai_2021}, OpenEarthMap~\cite{openearthmap_2023}, RescueNet~\cite{rescuenet_2023}, SkyScapes~\cite{skyscapes_2019}, UAVid~\cite{uavid_2020}, and iSAID~\cite{isaid_2019}. Global-LoveDA combines the EarthVQA and EarthVL-GLOBAL semantic subsets released in EarthVLSet, rather than the original LoveDA split. We retain the release names FLAIR-HUB, DOTA-v2, and FAIR1M-2.0 to distinguish dataset versions. Prepared sources may use subsets of the official collections: DOTA-v2 contains 2,248 tiles from 109 source images, while GID-15 and Five-Billion-Pixels share some source imagery. Thus, training counts do not represent unique original images. FLAIR-HUB uses 15 active CoSIA classes, and SkyScapes uses 19 classes from its Dense variant. Background and excluded-class policies follow the released evaluation catalogs rather than each benchmark's original protocol.

\paragraph{Table abbreviations.}
NWPU, PatternN., and FAIR1M denote NWPU-RESISC45, PatternNet, and FAIR1M-2.0. DeepG.-LC, FLAIR-H., Five-B.P., and G.-LoveDA denote DeepGlobe-LandCover, FLAIR-HUB, Five-Billion-Pixels, and Global-LoveDA. Potsdam, LandC.ai, OpenE.M., RescueN., and SkySc. denote ISPRS-Potsdam, LandCover.ai, OpenEarthMap, RescueNet, and SkyScapes. AeroSc. denotes AeroScapes. Abbreviations are used only for display; evaluation retains each dataset's full category catalog.

\subsection{Manually Annotated Source Test Sets}
We evaluate all methods against pixel-level GT on held-out samples from the 20 training sources. Table~\ref{tab:manual_semantic_tests} lists 1,000 manually annotated masks from the seven classification and detection datasets; the 13 pixel-labeled datasets use prepared held-out semantic test splits. Evaluation samples are excluded from adaptation by prepared sample ID. Potsdam uses a patch-level split in which training and test patches share parent orthophotos; its results therefore do not establish orthophoto-held-out spatial generalization. GID-15 uses an internal split of public annotations rather than the official hidden-label test set. We will release the evaluation samples and masks to support reproducibility.

\begin{table}[h]
\centering
\renewcommand{\arraystretch}{0.82}
\small
\setlength{\tabcolsep}{5pt}
\begin{tabular}{llrr}
\toprule
Dataset & Training label & $C_{\mathrm{train}}$ & Manual GT \\
\midrule
AID & Class & 30 & 90 \\
NWPU & Class & 45 & 135 \\
PatternN. & Class & 38 & 114 \\
\midrule
DIOR-R & Class + Box & 20 & 200 \\
DOTA-v2 & Class + Box & 18 & 160 \\
FAIR1M & Class + Box & 37 & 201 \\
xView & Class + Box & 60 & 100 \\
\midrule
Total & & & 1,000 \\
\bottomrule
\end{tabular}
\caption{Manual semantic test masks for seven coarsely annotated sources. Class indicates image-level labels; Class + Box includes bounding boxes. $C_{\mathrm{train}}$ counts queryable training categories, excluding background and ignore labels. Manual GT counts pixel-annotated images; these masks are not used for adaptation. NWPU, PatternN., and FAIR1M abbreviate NWPU-RESISC45, PatternNet, and FAIR1M-2.0 throughout.}
\label{tab:manual_semantic_tests}
\end{table}

\subsection{Training and Model Selection}
\label{sec:supp_evaluation_protocol}

\paragraph{Optimization and data exposure.}
We target 30 epochs with AdamW, learning rate $10^{-4}$, weight decay 0.05, polynomial decay power 0.9, and LoRA rank 8 at update scale 1.0. Four A100 GPUs each process four BF16 images. Each joint epoch samples 100,000 images at a Scene:Object:Pixel ratio of $1:2:4$, uniformly across datasets within each level, with up to 12 prompts per image. We validate every three epochs and at the final epoch, without using test masks for checkpoint selection. Augmentation applies random scaling in $[0.75,1.5]$, a $518\times518$ crop, flips, and right-angle rotations, with corresponding mask and box transforms. Source, ablation, and OOD evaluations use $518\times518$ windows at stride 388. Overlapping categorical predictions use center-weighted voting with a separable Hann window floored at 0.05 per axis. Small images are enlarged to the minimum window size, and predictions are restored to the GT grid.

\paragraph{Ablation exposure and completed budgets.}
Single-level runs sample 14,288 Scene, 28,576 Object, or 57,136 Pixel images per epoch, approximately matching the corresponding exposure in joint training. Scene~+~Object uses 42,864 samples per epoch at a $1:2$ ratio; Scene~+~Pixel uses 71,424 at approximately $1:4$; and Object~+~Pixel uses 85,712 at approximately $2:4$. Pairwise controls preserve per-level exposure. Unless explicitly ablated, all adapted variants share the HS-OVSeg architecture, preprocessing, and teacher-target construction. Comparisons use single runs; completed budgets are detailed below.

\paragraph{Spatial augmentation.}
Scale factors are sampled from the nominal $[0.75,1.5]$ range, subject to a minimum supporting a $518\times518$ crop. Labels undergo identical transforms, with nearest-neighbor interpolation for categorical masks. At inference, small images are enlarged while preserving aspect ratio, and predictions are mapped back to the original grid.

\paragraph{Model selection and completed training budgets.}
We set a 30-epoch budget and select checkpoints by mean validation mIoU across the 13 pixel-labeled datasets. Official validation splits are used when available; otherwise, randomly held-out training subsets serve as validation data and are excluded from adaptation. Test masks are used only for final evaluation.

\subsection{Evaluation Protocol}
\label{sec:supp_decoding}
All quantitative comparisons use single-label decoding: each pixel takes the highest-scoring fused-logit category, or background if the maximum is negative and background is defined. Presence probabilities do not reweight these scores. Independent prompt-wise masks instead threshold sigmoid probabilities at $0.5$; their IoUs are not directly comparable to single-label results. Direct pseudo-label evaluation in Sec.~\ref{sec:pseudo_label_quality} uses binary category masks consistent with training targets.

\paragraph{Valid pixels and averaging.}
Each dataset catalog defines queryable, background, excluded, and ignore categories. We accumulate dataset-level confusion counts before computing mIoU, averaging queryable classes with nonzero prediction--GT union. Background is excluded from the class mean, although valid background pixels may yield false positives; ignore pixels are excluded. Level means cover 3 Scene, 4 Object, or 13 Pixel datasets. The overall source mean directly averages all 20 datasets, while the five-dataset held-out mean is computed separately. Neither is an average of per-image IoUs.

\section{Additional Ablation Studies}
\label{sec:supp_target_ablation}

\subsection{Direct Pseudo-label Evaluation}
\label{sec:supp_pseudo_quality}
Table~\ref{tab:pseudo_quality} evaluates pseudo-label quality using the original scene labels and detection boxes for each manually annotated test image. We align category IDs through dataset-specific names and use source detection boxes, never boxes derived from manual masks. All annotated categories and instances are retained, without stochastic subsampling.

No random augmentation is applied. Small images are enlarged with preserved aspect ratio; larger images use overlapping $518\times518$ windows at stride 388, with center-weighted voting for independent binary targets. Boxes undergo corresponding scaling and clipping. Each tile follows the training target construction: normalized semantic responses for Scene and normalized semantic plus box-guided query evidence for Object. Padding is excluded from normalization, and manual GT validity is applied only during scoring. The threshold remains $0.5$ without test-mask tuning; binary targets return to the original grid by nearest-neighbor interpolation.

We accumulate per-category true positives, false positives, and false negatives across each dataset before computing mIoU, retaining overlapping predictions. Background can contribute false positives but is excluded from class averages; void pixels are ignored. Rejected Scene maps count as empty predictions rather than being discarded. Categories absent from coarse annotations also receive empty predictions, with any corresponding GT foreground counted as false negatives. This annotation-guided diagnostic measures target construction, not deployed box-free inference.

\paragraph{Interpreting teacher and student scores.}
Teacher evaluation uses annotation-guided independent binary targets; student evaluation uses the full vocabulary and single-label decoding without those annotations. Under these distinct protocols, Object target and student means are 30.83\% and 38.03\%, respectively. Together with the complementarity results, this supports learning from imperfect coarse supervision.

\subsection{Dataset-wise Ablation Results}
\label{sec:supp_full_ablation}

Tables~\ref{tab:supp_ablation_scene_object} and~\ref{tab:supp_ablation_pixel} report dataset-level mIoUs underlying main-text Table~\ref{tab:supervision_complementarity}. Native SAM3, the three single-level variants, the three pairwise variants, and HS-OVSeg correspond directly to the main table; additional rows cover target construction, prediction fusion, and text augmentation. All variants use identical fixed test subsets.

Scene means average AID, NWPU-RESISC45, and PatternNet; Object means average DIOR-R, DOTA-v2, FAIR1M-2.0, and xView; Pixel means cover the 13 datasets in Table~\ref{tab:supp_ablation_pixel}. All 20 means directly average the dataset scores, not the three level means. Aggregates precede display rounding. All pairwise controls evaluate the full vocabulary without image labels or boxes across all 20 datasets, using $518\times518$ windows at stride 388 and center-weighted voting as in Sec.~\ref{sec:supp_decoding}.

\paragraph{Cross-level transfer.}
Each single-level model improves the two untrained group means over native SAM3, though not necessarily every dataset. Without Pixel annotations, Scene~+~Object reaches 74.14\% Scene, 28.38\% Object, 25.03\% Pixel, and 33.07\% overall mIoU. Its Pixel performance indicates cross-level transfer, while the gap to the joint model's 64.13\% underscores the value of pixel GT. Joint training retains Pixel mIoU at 64.13\%, versus 64.07\% for Pixel-only, and improves Scene and Object means by 50.24 and 15.27 percentage points. Dataset-level gains and losses support aggregate preservation, not universal positive transfer or robustness to arbitrary noise. All comparisons are single runs under the budgets in Sec.~\ref{sec:supp_evaluation_protocol}.

\paragraph{Scene and Pixel supervision.}
Scene~+~Pixel achieves 76.03\% Scene, 24.43\% Object, 64.25\% Pixel, and 58.05\% overall mIoU. Scene improves by 0.22 points over Scene-only and 51.15 points over Pixel-only; Pixel remains at 64.25\%, compared with 64.07\% for Pixel-only. This variant leads the Scene and Pixel group means among supervision combinations, although its Object mean trails the all-level model's 38.03\%. Pixel scores include 73.01\% on DeepGlobe-LandCover, 60.42\% on FloodNet, and 46.25\% on RescueNet. Thus, combining scene-category coverage with dense labels does not recover the Object performance gained from object annotations.

\paragraph{Object and Pixel supervision.}
Object~+~Pixel achieves 27.59\% Scene, 38.82\% Object, 63.55\% Pixel, and 53.21\% overall mIoU. Its Object mean exceeds Object-only and Pixel-only by 8.73 and 16.06 points, while Pixel is 0.52 points below Pixel-only. Object scores reach 65.14\% on DIOR-R, 47.71\% on DOTA-v2, 22.97\% on FAIR1M-2.0, and 19.47\% on xView. Object mIoU is higher than HS-OVSeg's 38.03\%, at 38.82\%, but HS-OVSeg achieves higher Scene mIoU at 75.12\% versus 27.59\% and higher overall mIoU at 60.56\% versus 53.21\%. All-level supervision therefore yields the best overall balance, not the highest score for every group. Single runs limit attribution of small differences solely to supervision.

\begin{table*}[h]
\centering
\footnotesize
\setlength{\tabcolsep}{5pt}
\renewcommand{\arraystretch}{0.82}
\resizebox{\textwidth}{!}{%
\begin{tabular}{lrrrLrrrrL}
\toprule
Variant & AID & NWPU & PatternN. & Scene mean & DIOR-R & DOTA-v2 & FAIR1M & xView & Object mean \\
\midrule
Native SAM3 & 7.25 & 7.47 & 10.04 & 8.25 & 8.12 & 1.61 & 1.90 & 4.64 & 4.07 \\
\midrule
Scene only & 72.72 & \textbf{74.38} & \textbf{80.33} & 75.81 & 37.59 & 20.54 & 5.35 & 4.33 & 16.95 \\
Object only & 15.66 & 31.38 & 23.27 & 23.43 & 64.03 & 45.78 & 9.55 & 1.02 & 30.09 \\
Pixel only & 22.43 & 26.56 & 25.65 & 24.88 & 31.34 & 44.40 & 11.17 & 4.13 & 22.76 \\
Scene + Object & 71.99 & 71.22 & 79.22 & 74.14 & 62.32 & 39.90 & 9.24 & 2.06 & 28.38 \\
Scene + Pixel & 75.24 & 73.16 & 79.69 & \textbf{76.03} & 38.05 & 44.56 & 10.15 & 4.95 & 24.43 \\
Object + Pixel & 22.96 & 31.07 & 28.76 & 27.59 & \textbf{65.14} & 47.71 & \textbf{22.97} & \textbf{19.47} & \textbf{38.82} \\
\midrule
Scene presence only & 27.18 & 23.31 & 31.89 & 27.46 & 13.96 & 2.53 & 2.60 & 3.59 & 5.67 \\
Scene full-image labels & 52.43 & 41.04 & 52.05 & 48.51 & 7.33 & 3.28 & 0.75 & 0.72 & 3.02 \\
Object filled-box targets & 10.62 & 22.81 & 16.08 & 16.50 & 44.17 & 29.88 & 15.61 & 10.79 & 25.11 \\
Object teacher w/o box prompts & 23.27 & 28.62 & 27.32 & 26.40 & 41.97 & 23.33 & 12.60 & 3.23 & 20.28 \\
\midrule
Semantic input only & 75.32 & 70.40 & 79.23 & 74.98 & 63.82 & 47.01 & 22.26 & 16.42 & 37.38 \\
Query input only & \textbf{76.14} & 70.67 & 79.26 & 75.36 & 63.42 & 45.18 & 15.69 & 3.05 & 31.83 \\
Equal-weight fusion & 72.73 & 69.29 & 79.80 & 73.94 & 64.55 & 47.63 & \textbf{22.97} & 17.13 & 38.07 \\
Original class names only & 73.81 & 69.48 & 79.19 & 74.16 & 65.11 & 46.85 & 22.92 & 17.63 & 38.13 \\
\midrule
\textbf{HS-OVSeg} & 73.90 & 71.45 & 80.00 & 75.12 & 63.83 & \textbf{48.86} & 21.71 & 17.71 & 38.03 \\
\bottomrule
\end{tabular}}
\caption{Dataset-wise mIoU (\%) for three Scene and four Object test sources. All variants use identical fixed subsets and stride 388. Level means match main-text Table~\ref{tab:supervision_complementarity}; training budgets and ablation controls are described in Sec.~\ref{sec:supp_evaluation_protocol}.}
\label{tab:supp_ablation_scene_object}
\end{table*}

\begin{table*}[h]
\centering
\footnotesize
\setlength{\tabcolsep}{4pt}
\renewcommand{\arraystretch}{0.82}
\resizebox{\textwidth}{!}{%
\begin{tabular}{l*{13}{r}LM}
\toprule
Variant & \rotatebox{60}{DeepG.-LC} & \rotatebox{60}{FLAIR-H.} & \rotatebox{60}{Five-B.P.} & \rotatebox{60}{FloodNet} & \rotatebox{60}{GID-15} & \rotatebox{60}{G.-LoveDA} & \rotatebox{60}{Potsdam} & \rotatebox{60}{LandC.ai} & \rotatebox{60}{OpenE.M.} & \rotatebox{60}{RescueN.} & \rotatebox{60}{SkySc.} & \rotatebox{60}{UAVid} & \rotatebox{60}{iSAID} & \rotatebox{60}{Pixel mean} & \rotatebox{60}{All 20 mean} \\
\midrule
Native SAM3 & 12.42 & 12.13 & 6.31 & 17.61 & 5.99 & 25.09 & 33.90 & 30.56 & 27.26 & 6.43 & 20.52 & 34.16 & 10.05 & 18.65 & 14.17 \\
\midrule
Scene only & 41.04 & 18.53 & 8.26 & 20.77 & 3.23 & 25.11 & 20.86 & 27.66 & 31.59 & 14.26 & 6.36 & 30.45 & 27.38 & 21.19 & 28.54 \\
Object only & 28.67 & 18.03 & 10.03 & 10.36 & 1.02 & 9.80 & 30.35 & 12.40 & 30.44 & 10.40 & 15.45 & 35.14 & 46.09 & 19.86 & 22.44 \\
Pixel only & 72.92 & 63.66 & 63.44 & 60.13 & 48.99 & 63.70 & \textbf{82.88} & 83.49 & \textbf{66.78} & 44.09 & 43.30 & 76.94 & 62.62 & 64.07 & 49.93 \\
Scene + Object & 37.93 & 19.47 & 8.29 & 26.26 & 4.39 & 29.70 & 26.70 & 35.01 & 28.80 & 13.71 & 10.92 & 35.54 & 48.68 & 25.03 & 33.07 \\
Scene + Pixel & \textbf{73.01} & 63.28 & 65.51 & \textbf{60.42} & 48.66 & 64.02 & 82.82 & 83.21 & 66.39 & \textbf{46.25} & 43.76 & 76.58 & 61.30 & \textbf{64.25} & 58.05 \\
Object + Pixel & 72.55 & 63.69 & 63.64 & 60.21 & 48.54 & 63.61 & 82.33 & 83.56 & 65.89 & 42.18 & 43.35 & 75.33 & 61.20 & 63.55 & 53.21 \\
\midrule
Scene presence only & 15.47 & 10.66 & 11.34 & 30.98 & 8.00 & 30.42 & 48.25 & 45.91 & 21.96 & 12.77 & 15.38 & 33.84 & 8.49 & 22.58 & 19.93 \\
Scene full-image labels & 35.58 & 15.90 & 7.60 & 11.66 & 2.94 & 20.85 & 2.99 & 20.30 & 19.11 & 6.93 & 1.91 & 16.71 & 5.62 & 12.93 & 16.29 \\
Object filled-box targets & 22.05 & 15.39 & 6.74 & 13.18 & 0.60 & 14.63 & 15.86 & 34.26 & 21.62 & 9.77 & 8.15 & 23.70 & 39.46 & 17.34 & 18.77 \\
Object teacher w/o box prompts & 27.93 & 17.43 & 8.03 & 12.68 & 2.09 & 14.51 & 24.61 & 19.66 & 32.36 & 8.84 & 17.17 & 32.49 & 32.85 & 19.28 & 20.55 \\
\midrule
Semantic input only & 71.41 & 61.47 & 63.84 & 58.78 & 48.66 & 63.55 & 82.26 & 83.45 & 65.58 & 41.35 & 43.49 & 75.83 & \textbf{64.09} & 63.37 & 59.91 \\
Query input only & 72.24 & 63.41 & 62.37 & 58.97 & \textbf{49.36} & 63.55 & 82.67 & 83.58 & 65.71 & 35.86 & 42.47 & 75.97 & 59.79 & 62.77 & 58.47 \\
Equal-weight fusion & 72.09 & 63.91 & 63.07 & 59.40 & 48.62 & 63.77 & 82.75 & 83.36 & 66.30 & 41.02 & 44.17 & \textbf{76.95} & 62.18 & 63.66 & 60.08 \\
Original class names only & 72.59 & 64.08 & 62.89 & 59.72 & 48.73 & 61.95 & 82.58 & \textbf{83.64} & 66.32 & 44.27 & 43.91 & 76.51 & 63.54 & 63.90 & 60.29 \\
\midrule
\textbf{HS-OVSeg} & 71.96 & \textbf{64.61} & \textbf{65.62} & 60.17 & 48.64 & \textbf{65.12} & 81.94 & 83.54 & 66.05 & 42.54 & \textbf{44.48} & 76.58 & 62.49 & 64.13 & \textcolor{HSBestRed}{\textbf{60.56}} \\
\bottomrule
\end{tabular}}
\caption{Dataset-wise mIoU (\%) for 13 Pixel sources. Pixel mean averages these datasets; All 20 includes the seven sources in Table~\ref{tab:supp_ablation_scene_object}. Single-level and joint means match main-text Table~\ref{tab:supervision_complementarity}.}
\label{tab:supp_ablation_pixel}
\end{table*}

\begin{figure*}[h]
\centering
\includegraphics[width=\textwidth,height=0.95\textheight,keepaspectratio]{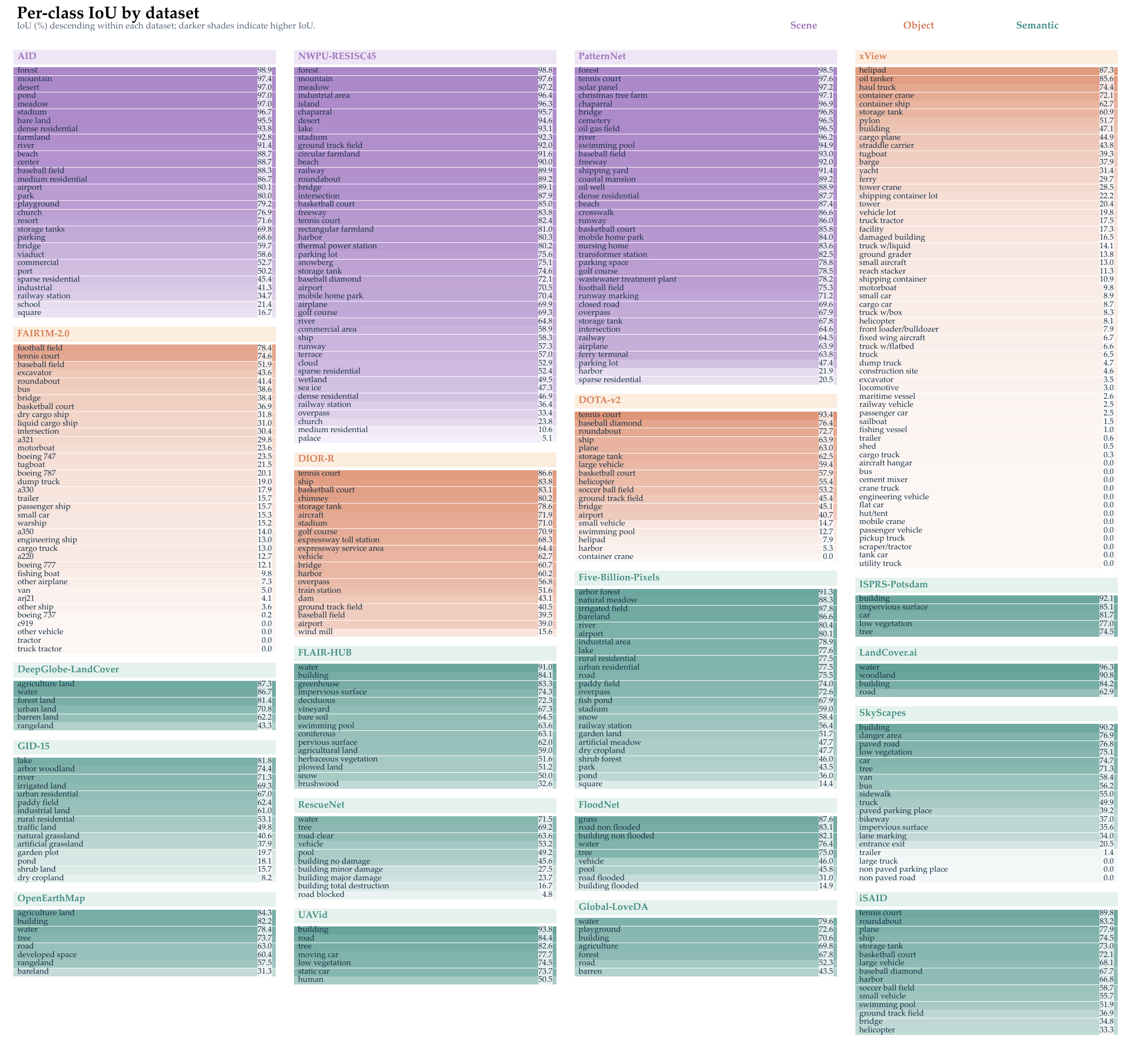}
\vspace{-0.8cm}
\caption{Class-wise IoU (\%) across 20 training-source test datasets using HS-OVSeg at stride 388. Classes are ranked by IoU within each dataset. Purple, orange, and green denote Scene, Object, and Pixel sources; darker shades indicate higher IoU.}
\label{fig:supp_class_iou}
\end{figure*}

\subsection{Class-wise Segmentation Performance}
\label{sec:supp_class_iou}

Figure~\ref{fig:supp_class_iou} reports class-wise IoUs for the HS-OVSeg checkpoint used in the main comparisons. Intersection and union are accumulated over each fixed test subset, excluding background and ignore classes. Categories are ranked within each dataset; identical names across datasets are not pooled.

\subsection{Scene-level Pseudo-label Construction}
Scene-only controls compare native SAM3, presence-only adaptation, full-image targets, and normalized online pseudo-labels on the same three datasets. Trained variants share the HS-OVSeg architecture, per-epoch Scene exposure, and 30 epochs. Presence-only omits pixel supervision; full-image targets assign the positive scene category to every valid pixel.

\begin{table}[h]
\centering
\renewcommand{\arraystretch}{0.82}
\small
\setlength{\tabcolsep}{8pt}
\resizebox{\columnwidth}{!}{%
\begin{tabular}{lrrrM}
\toprule
Scene strategy & AID & NWPU & PatternN. & Mean \\
\midrule
Native SAM3 & 7.25 & 7.47 & 10.04 & 8.25 \\
Presence only & 27.18 & 23.31 & 31.89 & 27.46 \\
Full-image labels & 52.43 & 41.04 & 52.05 & 48.51 \\
Normalized pseudo-labels & \textbf{72.72} & \textbf{74.38} & \textbf{80.33} & \textbf{75.81} \\
\bottomrule
\end{tabular}%
}
\caption{Scene-only ablation in semantic mIoU (\%). NWPU and PatternN. denote NWPU-RESISC45 and PatternNet. Native SAM3 is unadapted.}
\label{tab:scene_ablation}
\end{table}

\paragraph{Analysis.} Normalized pseudo-labels achieve 75.81\% Scene mIoU, compared with 48.51\% for full-image labels and 27.46\% for presence-only adaptation, supporting spatially selective pixel supervision.

\subsection{Object-level Pseudo-label Construction}
Table~\ref{tab:object_ablation} compares object-only controls trained on the same four detection sources and evaluated at epoch 30. Filled-box targets use box interiors instead of teacher masks. The no-box-prompt variant removes geometric prompts but retains annotated boxes for query association. The reference combines normalized teacher semantic probabilities with selected, unnormalized query-mask probabilities. Student fusion-input controls appear separately in Table~\ref{tab:pixel_fusion_ablation}.

\begin{table*}[h]
\centering
\renewcommand{\arraystretch}{0.8}
\small
\setlength{\tabcolsep}{9pt}
\begin{tabular}{lrrrrM}
\toprule
Object strategy & DIOR-R & DOTA-v2 & FAIR1M & xView & Mean \\
\midrule
Native SAM3 & 8.12 & 1.61 & 1.90 & 4.64 & 4.07 \\
Filled-box targets & 44.17 & 29.88 & \textbf{15.61} & \textbf{10.79} & 25.11 \\
Teacher without box prompts & 41.97 & 23.33 & 12.60 & 3.23 & 20.28 \\
Normalized Semantic + Query & \textbf{64.03} & \textbf{45.78} & 9.55 & 1.02 & \textbf{30.09} \\
\bottomrule
\end{tabular}
\caption{Object-only ablation in semantic mIoU (\%), not detection or instance metrics. FAIR1M denotes FAIR1M-2.0; all trained controls use 30 epochs.}
\label{tab:object_ablation}
\end{table*}

\paragraph{Analysis.} Object-only training reaches 30.09\% Object mIoU, versus 25.11\% with filled boxes and 20.28\% without teacher box prompts. These controls assess target construction, whereas fusion-input controls assess student prediction pathways.

\subsection{Integrating Semantic and Query Evidence}
\label{sec:supp_fusion_ablation}

Table~\ref{tab:pixel_fusion_ablation} compares learned local fusion with Semantic-only, Query-only, and calibrated equal-weight fusion using identical training sources and pseudo-labels. Single-input variants retain the convolutional head but zero the other input to isolate each pathway.

\begin{table}[h]
\centering
\footnotesize
\setlength{\tabcolsep}{6pt}
\renewcommand{\arraystretch}{0.82}
\resizebox{\columnwidth}{!}{%
\begin{tabular}{lLLLM}
\toprule
Variant & Scene & Object & Pixel & All mean \\
\midrule
Semantic input only & 74.98 & 37.38 & 63.37 & 59.91 \\
Query input only & \textbf{75.36} & 31.83 & 62.77 & 58.47 \\
Equal-weight fusion & 73.94 & \textbf{38.07} & 63.66 & 60.08 \\
\textbf{HS-OVSeg} & 75.12 & 38.03 & \textbf{64.13} & \textcolor{HSBestRed}{\textbf{60.56}} \\
\bottomrule
\end{tabular}%
}
\caption{Semantic--Query fusion ablation across 20 training-source datasets in mIoU (\%). Equal-weight fusion averages calibrated logits.}
\label{tab:pixel_fusion_ablation}
\end{table}

Combining Semantic and Query predictions improves overall mIoU over either input alone. HS-OVSeg reaches 60.56\%, exceeding equal-weight fusion by 0.48 percentage points, with competitive results across all three levels.

\subsection{Class-Text Augmentation}
The original-class-name control and HS-OVSeg share all 20 training sources, normalized teacher targets, and the presence-plus-fused-segmentation objective. Without class-text augmentation, the control reaches 60.29\% overall mIoU, versus 60.56\% for HS-OVSeg. Both use identical overlapping inference at stride 388, isolating class-text augmentation for these checkpoints.

\begin{table*}[t]
\centering
\renewcommand{\arraystretch}{0.82}
\small
\setlength{\tabcolsep}{8pt}
\begin{tabular}{lLLLM}
\toprule
Variant & Scene mIoU & Object mIoU & Pixel mIoU & All 20 mIoU \\
\midrule
Original class names only & 74.16 & \textbf{38.13} & 63.90 & 60.29 \\
\textbf{HS-OVSeg} & \textbf{75.12} & 38.03 & \textbf{64.13} & \textcolor{HSBestRed}{\textbf{60.56}} \\
\bottomrule
\end{tabular}
\caption{Class-text augmentation across 20 sources, with 26 epochs per run. Scores are mIoU (\%).}
\label{tab:optimization_ablation}
\end{table*}

\section{Comparison with Existing Methods}
\label{sec:additional_analyses}

\subsection{Compared Methods and Pretraining Overlap}
We compare against four general-purpose methods---native SAM3, CAT-Seg, CorrCLIP, and Trident---and nine RS-oriented methods: RemoteCLIP, RSKT-Seg, GSNet, LDSeg, SegEarth-OV, GeoRSCLIP, RSCLIP, SegEarth-OV3, and SkySense-O. All use released weights and method-specific preprocessing on identical evaluation images and category catalogs. RemoteCLIP and GeoRSCLIP use our dense prediction adapters, not author-native segmentation heads. Some released RS models may have pretrained on overlapping dataset families, including iSAID, Potsdam, or LoveDA. Their scores reflect released-checkpoint transfer, not training under the HS-OVSeg data budget. The five held-out datasets are excluded from HS-OVSeg adaptation, but not necessarily from every baseline's upstream pretraining.

\subsection{Level-wise Comparison Summary}
\label{sec:supp_dataset_comparison}
Table~\ref{tab:source_comparison} summarizes performance across levels. HS-OVSeg achieves 75.12\% Scene, 38.03\% Object, and 64.13\% Pixel mIoU, outperforming the strongest overall baseline, SkySense-O, in each group. This supports unified segmentation across scene concepts, object entities, and land-cover categories.

\begin{table}[h]
\centering
\renewcommand{\arraystretch}{0.82}
\footnotesize
\setlength{\tabcolsep}{10pt}
\resizebox{\columnwidth}{!}{%
\begin{tabular}{lLLLM}
\toprule
 & \multicolumn{4}{c}{mIoU (\%) $\uparrow$} \\
\cmidrule(lr){2-5}
Method & Scene & Object & Pixel & All 20 mean \\
\midrule
Native SAM3~\cite{sam3_2026} & 8.25 & 4.07 & 18.65 & 14.17 \\
RemoteCLIP~\cite{remoteclip_2024} & 19.16 & 6.63 & 18.48 & 16.21 \\
RSKT-Seg~\cite{rskt_seg_2026} & 19.66 & 3.57 & 20.64 & 17.08 \\
GSNet~\cite{gsnet_2025} & 13.69 & 3.96 & 22.19 & 17.27 \\
LDSeg~\cite{ldseg_2026} & 13.77 & 4.22 & 24.87 & 19.07 \\
SegEarth-OV~\cite{segearth_ov_2025} & 12.69 & 5.05 & 25.47 & 19.47 \\
RSCLIP~\cite{rsclip_2026} & 13.11 & 5.06 & 25.71 & 19.69 \\
GeoRSCLIP~\cite{georsclip_2024} & 18.53 & 6.69 & 24.02 & 19.73 \\
Trident~\cite{trident_2025} & 16.62 & 4.92 & 26.39 & 20.63 \\
CorrCLIP~\cite{corrclip_2025} & 20.30 & 6.73 & 25.49 & 20.96 \\
SegEarth-OV3~\cite{segearth_ov3_2025} & 20.35 & 11.72 & 26.93 & 22.90 \\
CAT-Seg~\cite{cat_seg_2024} & 20.76 & 5.51 & 29.28 & 23.24 \\
SkySense-O~\cite{skysense_o_2025} & 33.08 & 23.56 & 31.44 & 30.11 \\
\midrule
\textbf{HS-OVSeg} & \textbf{75.12} & \textbf{38.03} & \textbf{64.13} & \textcolor{HSBestRed}{\textbf{60.56}} \\
\bottomrule
\end{tabular}%
}
\caption{Level-wise summary of main-text Table~\ref{tab:source_comparison_per_dataset}, ordered by All 20 mean. Group means average 3 Scene, 4 Object, and 13 Pixel datasets; All 20 weights datasets equally.}
\label{tab:source_comparison}
\end{table}

\subsection{Qualitative Comparison across Annotation Levels}
\begin{figure*}[t]
\centering
\includegraphics[width=\textwidth,height=0.93\textheight,keepaspectratio]{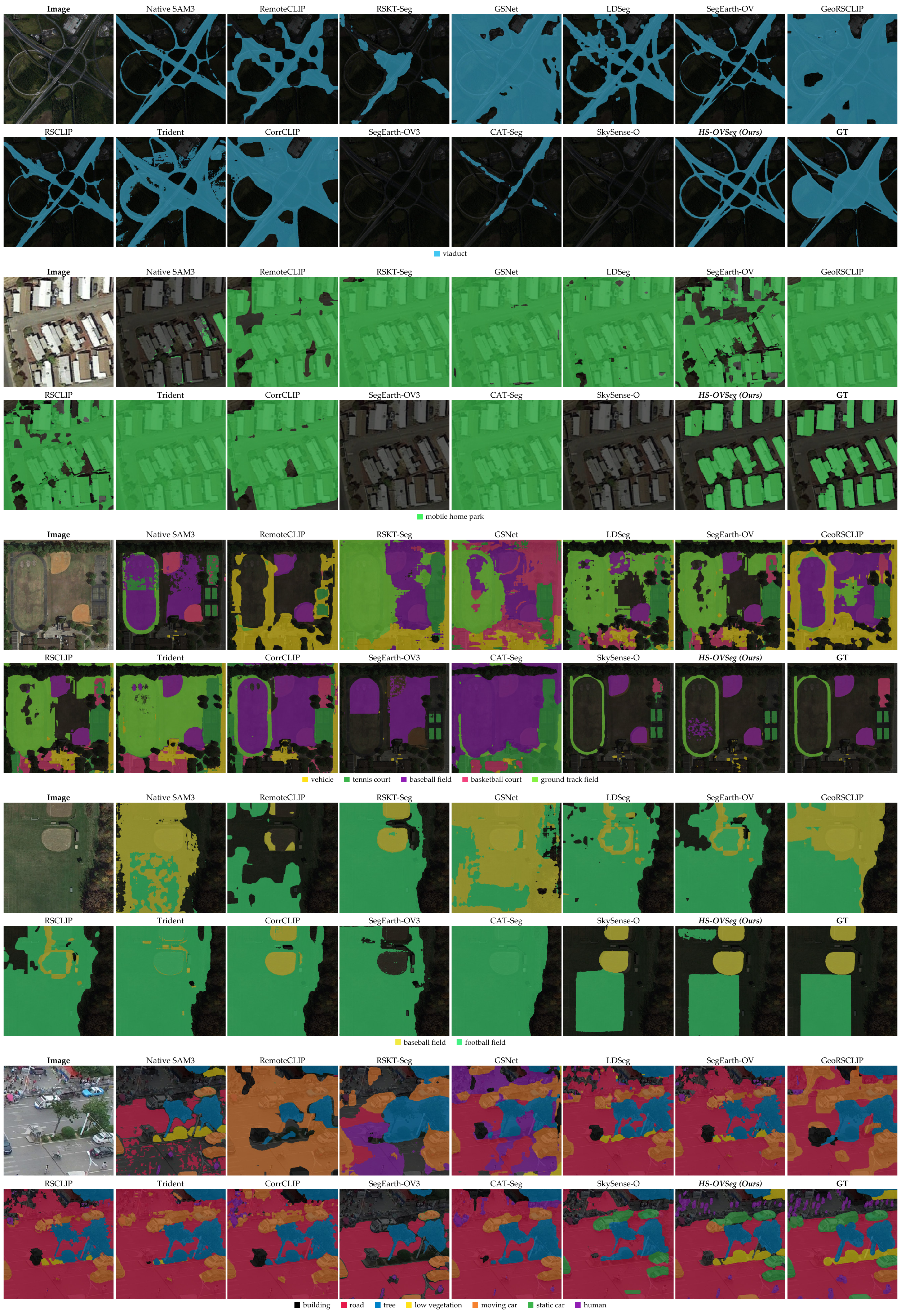}
\caption{Additional qualitative comparisons.}
\label{fig:supp_comparison_samples}
\end{figure*}

Figure~\ref{fig:supp_comparison_samples} provides additional method comparisons with reference masks and category legends.

Main-text Fig.~\ref{fig:qualitative_comparison} shows an AID airport, a DOTA-v2 sports facility, and an OpenEarthMap tile. Scene and Object references are manually annotated. All methods receive only GT-present category prompts, without boxes or GT pixel locations. The Scene legend is omitted; Object and Pixel legends list reference-present classes. All predictions use overlapping $518\times518$ windows at stride 388, as in quantitative evaluation. HS-OVSeg applies a foreground threshold of $0.5$, and predictions and GT are overlaid on the same RGB image. Unlike quantitative evaluation, these illustrations use GT-present rather than full-vocabulary prompts; conclusions rely on the full test results.

\subsection{Parameter and Computational Efficiency}
HS-OVSeg trains 9,558,617 parameters, including LoRA adapters and 453 local fusion-head parameters. Under the matched RTX 3090 protocol in Table~\ref{tab:adaptation_cost}, HS-OVSeg takes 161.9 ms and 4.37 GB peak memory, versus 113.2 ms and 4.33 GB for frozen SAM3.

\begin{table*}[h]
\centering
\renewcommand{\arraystretch}{0.82}
\footnotesize
\setlength{\tabcolsep}{14pt}
\resizebox{\textwidth}{!}{%
\begin{tabular}{lrrrrMM}
\toprule
Adaptation & Epochs & Trainable (M) & Infer. mem. (GB) & Infer. (ms) & Source mIoU & Held-out mIoU \\
\midrule
Frozen SAM3 & 0 & 0 & 4.33 & 113.2 & 14.17 & 29.42 \\
HS-OVSeg & 26 & 9.559 & 4.37 & 161.9 & \textcolor{HSBestRed}{\textbf{60.56}} & \textcolor{HSBestRed}{\textbf{43.33}} \\
\bottomrule
\end{tabular}
}
\caption{Parameter efficiency and segmentation performance. Sequential RTX 3090 inference uses one $518\times518$ image, eight prompts, BF16, five warmup and 30 timed iterations; image transfer and disk I/O are excluded. Memory is peak allocated inference memory in decimal GB. Source and held-out means weight 20 and five datasets equally.}
\label{tab:adaptation_cost}
\end{table*}

\section{Out-of-Domain Evaluation}
\label{sec:supp_ood_evaluation}

\subsection{Datasets and Fixed Subsets}
We evaluate DLRSD~\cite{dlrsd_2018}, AeroScapes~\cite{aeroscapes_2018}, WHDLD~\cite{dlrsd_2018,whdld_2020}, VDD~\cite{vdd_2025}, and DynamicEarthNet without target-dataset adaptation; Table~\ref{tab:ood_datasets} summarizes them. All provide pixel GT and are evaluated as semantic segmentation, despite differing emphasis on scenes, objects, or land cover. This measures cross-dataset transfer, not necessarily unseen-category generalization: held-out categories may overlap the adaptation vocabulary.

\begin{table}[h]
\centering
\renewcommand{\arraystretch}{0.82}
\small
\setlength{\tabcolsep}{4pt}
\resizebox{\columnwidth}{!}{%
\begin{tabular}{llrrr}
\toprule
Dataset & Emphasis & Classes & Samples & Size (px) \\
\midrule
DLRSD & Scene-related regions & 17 & 2,000 & $256\times256$ \\
AeroSc. & Objects / land cover & 11 & 2,000 & $518\times518$ \\
WHDLD & Land cover & 6 & 2,000 & $256\times256$ \\
VDD & Urban semantics & 7 & 40 & $4000\times3000$ \\
DynamicEarthNet & Land cover & 7 & 176 & $518\times518$ \\
\bottomrule
\end{tabular}%
}
\caption{Five held-out evaluation datasets.}
\label{tab:ood_datasets}
\end{table}

\paragraph{Fixed evaluation subsets.}
For test splits larger than 2,000 samples, we select at most 2,000 fixed samples to preserve source coverage and semantic diversity, using provenance and GT class histograms rather than predictions. Large AeroScapes images are tiled before selection. DynamicEarthNet uses 176 non-overlapping tiles from 44 AOIs acquired on July 1, 2018, at native 3 m resolution; padding is ignored. All methods share sample lists and category catalogs. Results refer to these subsets, not the full original test sets.

\subsection{Cross-Dataset Transfer}
HS-OVSeg leads the five-dataset average at 43.33\%, but not every dataset. SegEarth-OV3 leads AeroScapes at 52.22\%, and SkySense-O leads DynamicEarthNet at 36.50\%. HS-OVSeg achieves 45.39\% on DLRSD, 43.96\% on WHDLD, and 52.65\% on VDD. These results support cross-dataset transfer with dataset-dependent variation. Held-out datasets are excluded from HS-OVSeg adaptation, not necessarily from baselines' upstream pretraining; queried categories may also have been seen.

\section{Multi-granularity Prediction}
\label{sec:supp_multigranularity}

\subsection{Prompt-wise Visualization and Benchmark Decoding}
Figure~\ref{fig:multi_granularity_prediction} shows large-area predictions under separate Scene, Object, and Pixel vocabularies. Independent Scene and Object masks may overlap, whereas benchmarks use the single-label decoding in Sec.~\ref{sec:supp_decoding}. Visualization thresholds and overlays do not affect reported mIoUs. Figure~\ref{fig:qualitative_comparison} uses single-label decoding with GT-present prompts; benchmarks query the full vocabulary.

\section{Implementation Details of HS-OVSeg}
\label{sec:supp_implementation}

\subsection{Model Adaptation}
Eligible linear and convolutional weights receive low-rank updates; biases and one-dimensional normalization parameters remain frozen. Each tensor has independent factors with effective rank $\min(8,d_{\mathrm{out}},d_{\mathrm{in}})$ and update scale one, without rank division. An adapter receives gradients only from objectives along its computation path. Presence and fused segmentation losses update their respective paths, including both prediction branches and shared decoder components. No instance-matching loss is used. Only the frozen teacher receives geometry prompts. The trainable student fusion head propagates segmentation gradients through both inputs.

\subsection{Online Pseudo-label Generation}
The teacher uses the frozen pretrained weights with LoRA disabled and stop-gradient; it receives no exponential-moving-average updates. Teacher and student share augmented images and category text, while the object teacher also receives annotated boxes. Pseudo-labels are regenerated for each sampled view, not cached offline. Semantic logits are resized bilinearly before sigmoid; object query masks undergo sigmoid and maximum aggregation before resizing, as in Eq.~\eqref{eq:object_aggregation}. GT masks retain discrete labels after spatial transforms. Pixel samples bypass the teacher; Scene and Object samples require teacher inference.

\subsection{Category and Instance Sampling}
Each image uses up to $12$ prompts from its dataset vocabulary. Positives come from scene labels, object categories retained after augmentation, or valid semantic masks. If positives exceed the budget, a seed-dependent subset is chosen; remaining slots use annotation-absent categories without a fixed positive-to-negative ratio. Thus, per-image CE does not use a global cross-dataset category catalog.

Object batches retain at most $36$ boxes, first distributing them across sampled image--category groups and then filling the remaining budget. Categories excluded by instance sampling are omitted from prompts rather than treated as negatives; omitted boxes for retained categories are excluded from pixel supervision. Negative sampling prioritizes categories seen in other images from the same dataset and batch, using at most half the negative budget rounded upward when available. Remaining negatives come from annotation-absent categories, which may still be visually present if annotations are incomplete.

\subsection{Query-mask Stability and Aggregation}
\label{sec:supp_query_stability}
We compute query-mask stability from pre-sigmoid logits $m_{c,j}$, with $M_{c,j}=\sigma(m_{c,j})$ as in Eq.~\eqref{eq:sam3_outputs}. On the native mask grid, define
\begin{equation}
\begin{aligned}
\mathcal{A}_{c,j}^{+}&=\{u:m_{c,j}(u)>\delta\},\\
\mathcal{A}_{c,j}^{-}&=\{u:m_{c,j}(u)>-\delta\},\\
\kappa_{c,j}&=
\begin{cases}
\dfrac{|\mathcal{A}_{c,j}^{+}|}{|\mathcal{A}_{c,j}^{-}|},
& |\mathcal{A}_{c,j}^{-}|>0,\\
0,&\text{otherwise},
\end{cases}
\end{aligned}
\label{eq:mask_stability}
\end{equation}
Here $\delta=0.2$ in logit space corresponds to probability thresholds near $0.550$ and $0.450$. Since $\mathcal{A}_{c,j}^{+}\subseteq\mathcal{A}_{c,j}^{-}$, their area ratio equals binary-mask IoU. Higher stability indicates less foreground change under threshold perturbation; empty unions score zero.

Stability is computed before resizing without learned parameters. Queries are ranked and weighted by $\kappa_{c,j}q_{c,j}$; the top eight are retained. Pixel-wise maximum aggregation precedes bilinear resizing, as in Eq.~\eqref{eq:student_query_aggregation}.

\subsection{Unified Losses}
Following Eq.~\eqref{eq:pixel_fusion}, let $Z_c\in\mathbb{R}^{H\times W}$ be the fused logit map and $F_c=\sigma(Z_c)\in[0,1]^{H\times W}$ the foreground probability for category $c$. Let $v_c(u)\in\{0,1\}$ mark valid supervision at pixel $u$, with $n_c=\sum_u v_c(u)$. For $n_c>0$,
\begin{equation}
\begin{aligned}
L_c^{\mathrm{BCE}}
&=\frac{1}{n_c}\sum_u v_c(u)\,
\operatorname{BCEWithLogits}(Z_c(u),Y_c(u)),\\
L_c^{\mathrm{Dice}}
&=1-\frac{2\sum_u v_c(u)F_c(u)Y_c(u)+1}
{\sum_u v_c(u)[F_c(u)+Y_c(u)]+1}.
\end{aligned}
\label{eq:supp_binary_losses}
\end{equation}
Both terms are averaged over category maps with valid minibatch supervision, excluding padding and invalid pixels. Accepted Scene positives use all valid pixels; Scene negatives use only the union of accepted positive regions. Object targets exclude omitted same-category boxes, and Pixel targets use the GT validity mask.

For each image, let $\mathcal{U}_I$ denote pixels with exactly one valid positive sampled target, with category $t(u)$. Multiclass CE is
\begin{equation}
L_I^{\mathrm{CE}}
=-\frac{1}{|\mathcal{U}_I|}\sum_{u\in\mathcal{U}_I}
\log\frac{\exp Z_{t(u)}(u)}
{\sum_{c\in\mathcal{C}_I}\exp Z_c(u)}.
\label{eq:supp_multiclass_ce}
\end{equation}
CE averages images with nonempty $\mathcal{U}_I$. All sampled categories, including negatives, enter the softmax denominator. Pixels with overlapping or absent positives are excluded only from CE and may still contribute valid BCE and Dice targets. Presence BCE averages sampled image--category pairs. Pixel losses are zero without eligible supervision. All four loss weights equal one, and each batch contains one annotation granularity.

\subsection{Normalization and Spatial Guidance}
Both pseudo-label branches use valid-pixel quantiles of $2\%$ and $98\%$, then threshold at $0.5$. Scene normalization sets $\epsilon_s=10^{-8}$ and rejects positive maps with fewer than two valid pixels, non-finite quantiles, or a quantile span at most $10^{-8}$. Such maps provide no positive pixel supervision, but presence targets remain unchanged. Object normalization uses $\epsilon_o=10^{-6}$ as a denominator floor while retaining query evidence. A constant object semantic map becomes zero, leaving the query branch available. Neither branch applies an additional absolute-response or presence-based quality filter; normalization may amplify weak nondegenerate responses. Normalized and fused maps are target-construction scores, not calibrated foreground probabilities.

\paragraph{Box geometry.}
Each retained box is supplied separately as a positive geometric prompt. Prompts are normalized axis-aligned boxes; rotated annotations use enclosing axis-aligned boxes after augmentation. Query selection equally weights predicted-box IoU and sigmoid class-matching score. Same-category responses are merged by pixel-wise maximum. Boxes guide localization without hard masking, so pseudo-foreground may extend beyond them.

\end{document}